\documentclass[conference]{IEEEtran}

\usepackage[pdftex]{graphicx}
\usepackage[cmex10]{amsmath}
\usepackage{algorithm}
\usepackage{algpseudocode}
\usepackage[caption=false,font=footnotesize]{subfig}
\usepackage{xcolor}
\usepackage{url}

\title{Building an Affordances Map with Interactive Perception}

\author{
\IEEEauthorblockN{
Leni K. Le Goff, Oussama Yaakoubi,  Alexandre Coninx and Stephane Doncieux}}

\begin{document}

\maketitle

\begin{abstract}
Robots need to understand their environment to perform their task. If it is possible to pre-program a visual scene analysis process in closed environments, robots operating in an open environment would benefit from the ability to learn it through their interaction with their environment. This ability furthermore opens the way to the acquisition of affordances maps in which the action capabilities of the robot structure its visual scene understanding. We propose an approach to build such affordances maps by relying on an interactive perception approach and an online classification. In the proposed formalization of affordances, actions and effects are related to visual features, not objects, and they can be combined. We have tested the approach on three action primitives and on a real PR2 robot. 
\end{abstract}

\begin{IEEEkeywords}
Autonomous exploration, Affordance learning, Interactive Perception, Perceptual Map
\end{IEEEkeywords}

\section{Introduction}\label{sec:intro}

Nowadays, robots can achieve specific tasks with a high accuracy in controlled environments, such as automated factories. In such environments, the engineers can anticipate all the aspect of the problem at hand and then simply program the robot to achieve its goal. However, in open and dynamic environments, it is difficult to anticipate everything. To solve tasks in such a context, robots need adaptive skills. A way to approach this issue is to let explores its surrounding and learn from its experiences. By exploring its environment, the robot is able to build its own representation thereof according to its embodiment, skills and goals.

The psychologist E. Gibson claimed that acquiring perception is "discovering distinctive features and invariant properties of things and events" \cite{gibson2000perceptual} and "discovering the information that specifies an affordance." \cite{gibson2003world}. In other words, the system must isolate regularities and invariance in the data collected during an exploration to build representations. And, these representations are affordances. The concept of affordances was introduced by J.J. Gibson \cite{gibson1966senses,gibson1979}.


With this concept, J.J. Gibson wanted to highlight that objects have inherent "values" and "meanings" which could be perceived by an agent and could be linked to its possible actions on those objects. An animal or a human thus perceives the world through the actions it can perform according to their abilities and the elements in the environment.  J.J. Gibson's original definition of affordances was then refined and clarified by further work from the ecological psychology community, and several main conclusions were drawn:
\begin{itemize}
\item Affordances \textit{emerge} from the relation between the agent and the environment \cite{chemero2003outline};
\item \textit{Functionality} is an inherent property of objects or parts of the environment. A functionality could become an affordance if the agent has some knowledge about it and if the agent is able to use it \cite{steedman2002formalizing,steedman2002plans};
\item Affordances are not always self-evident. Therefore learning and exploration could be needed to perceive affordances. \textit{Signifiers} could be built to help an agent perceive affordances \cite{norman2013design}.  
\end{itemize}

In this paper, we state that an affordance is an emergent relationship in the agent-environment system. Thus, an affordance is a relationship between a sensory signal, the agent skills and the possible effect that would result from the agent's actions. Affordances are learned from experience of the agent interacting with the real world, and as a result of this learning, affordances can then be directly perceived in the environment. Moreover, for the affordances to be learned, the environment needs to have distinctive and coherent sensory signals associated with actions and effects, in other words, they need to be discoverable.

The work, presented in this paper, proposed a system to learn a perceptual representation based on affordances.  The aim is to answer to the following problematic: \textit{How can a robot with a toolbox of motor primitives build a representation of the environment based on affordances by autonomous exploration ?} The robotic system learns from data collected during an autonomous exploration by interacting with the environment. This approach follows the interactive perception paradigm.
  
Interactive perception aims at learning perception from interaction. According to J. Bohg et al. \cite{bohg2017interactive}, a robotic system, with interactive perception, isolates regularities in the combined space of sensory signals, motor commands and time. This meets the vision of E. Gibson about learning. Therefore, it is natural to use interactive perception to let a system autonomously learn affordances. 

However, most works in interactive perception are interested to learn objects representations for recognition, segmentation or manipulation. To achieve their goal, these methods  need to introduce assumptions about the structure of the environment or about the objects themselves. These assumptions reduce the range of environments that the robot could face. One of our previous works \cite{legoff2019boostrapping} addresses this issue by proposing a method to learn a perceptual map, called \textit{relevance map}, through interactive perception with minimum environment-specific assumptions. 

This paper presents an extension of this previous work. In our previous work, a relevance map is built based on data collected thanks to the interaction of a robot with an environment through a push primitive. This approach is within the scope of interactive perception as it learns a representation of the world through interactions with an environment. This relevance map was representing the relevant areas in a visual scene for the push primitive, i.e. the areas that would produce an effect after the application of the push primitive. Thus, the relevance map represents areas which afford a certain action. In the present study, relevance maps relative to several affordances are learned: pushable objects, liftable objects, and activable push-button. These maps are then combined to produce an affordances map. 
 This affordances map is a starting point for further developmental steps, and provide the knowledge needed to bootstrap a decision process \cite{doncieux2018framework}.

The main contribution of this work is a modular framework to learn low level affordances represented by a perceptual map. The affordances map gives to the robot a rich and direct perception of its surrounding according to the actions it can perform. This is close to Gibson's first conception of affordances.

This paper is organized as follows : Related works about affordance learning are described in section \ref{sec:rela_works}, then the proposed method is explained in section \ref{sec:method}, experiments and results are presented in sections \ref{sec:exp} and \ref{sec:res} and to conclude a discussion is proposed in section \ref{sec:disc}.

\section{Related Works}\label{sec:rela_works}

Affordances have raised a lot of interest in the developmental robotics community these last ten years, as shown by the numerous reviews and surveys dedicated to this topic \cite{Sahin2007,horton2012affordances,min2016affordance,jamone2018affordances,zech2017computational}.

According to a recent survey \cite{zech2017computational}, among 146 reviewed papers, 104 papers consider learning affordances directly from a meso level, i.e. considering objects as a whole, while only 27 papers consider it from global level, i.e. by considering the whole environment and only 15 papers from a local level. 
With the global level, considering the whole environment allows the learning system to integrate the context. The context is important to predict or to do recognition of high-level affordances. 
Most papers on affordance use the meso level because for most actions having a complete model of an object is practical. For instance, for successful grasps, the object states such as orientation and position or shape are important information. Learning affordances at a local level allows the system to perceive them directly, which is in line with Gibson's view. Moreover considering the local level is simpler and is thus suitable to bootstrap the system. 

The proposed method is based on learning affordances from local visual features, so from the local level. Therefore, this section is focused on papers interested in learning affordances at a local level. 
 From these 15 papers, 11 are interested in linking local descriptors to the possible actions applicable in the present environment for quick or direct perception of affordances. From these papers, 6 learn from exploration using an interactive perception approach. This shows that the question of learning affordances from local features using exploration has not been extensively studied yet. This section reviews different groups of works addressing this question. A first group aims at learning several kinds of affordances with supervised learning on annotated datasets, a second one focuses on the object grasping issue and finally, works that do not fit in these two categories are mentioned.

Some studies use an annotated dataset to train a model of affordance classification and then integrate this model in a robotic framework, as a tool for planning, task solving or manipulation. Myers et al. \cite{myers2015affordance} study tool use affordances. They train a classifier on superpixels using SLIC. Ashanta et al. \cite{Achanta2010} have extended it to work on RGB-D images, with features related to shape. Two classifiers are proposed in this work. A first one is called superpixel hierarchical matching which is computationally demanding and slow for prediction. The second one is a structured random forest which achives fast prediction and is therefore suited to real-time systems, but this last classifier is trained offline. AfRob method proposed by Varadarajan and Vincze \cite{varadarajan2012afrob} is used to classify affordances from 2D images. It is a deep neural network trained in batch. AfRob is the adaptation of, previously proposed, AfNet, from the same authors, to robotics constraints (fast prediction, light computation). Katz et al. \cite{katz2014perceiving} aim at detecting affordances from stacks of objects. With this aim, an SVM linear classifier is used to learn pulling, pushing and grasping affordances. As they use objects with simple shapes and only consider their facets as features, i.e. small planar surfaces which compose a 3D shape, they can use a simple linear classifier, especially if trained offline on an annotated dataset. In the same idea,  Kim and Sukhatme \cite{kim2014semantic} proposed a method to detect affordances of surfaces based on a geometrical analysis of the pointcloud, K-means clustering, and logistic regression. 

Those methods proposed efficient tools for robotic systems to detect affordances, but they are all based on supervised learning on datasets annotated by a human expert. Annotating is a costly process that naturally limits the learned model to the datasets produced by the expert. Moreover, affordances in ecological psychology depend on the agent body structure and on the actions it is capable of. Another approach is thus to let the robot explores its environment with one or several actions and collects information about the affordances in its surrounding and discovers by itself the affordances.

A group of works \cite{dang2014semantic,bierbaum2009grasp,montesano2009learning,Popovic2011,kruger2011grasp,kraft2010development} are focused on building affordance maps of successful grasp on an object. Bierbaum et al. \cite{bierbaum2009grasp} let a robotic hand with tactile sensors explore an unknown object in simulation. The robot hand has five fingers including a thumb. The system detects a potential grasp by finding opposite flat surfaces. Then, candidate areas for grasping are determined offline on the basis of the geometrical analysis of local shape features. The analysis is a heuristic based on the configuration of the hand used. Alternatively, Montesano and Lopes \cite{montesano2009learning} propose a trial and error process to determine the probability of success of a grasp on parts of an object. Learning is based on local visual features in a Bayesian framework. The robot tries to grasp several times the same object part and, with a Bernoulli-beta distribution based on the successes or failures, the system determines the probability of the graspability of this part.
In the same idea, Dang and Allen \cite{dang2014semantic} proposed a system that learns a graspable affordance map on objects but they add what they call semantic constraints. These constraints are designed by a human to force grasping to be compatible with a specific task. 
In the same way, Popovic et al. \cite{Popovic2011,kruger2011grasp} use  Early Cognitive Vision (ECV, \cite{kruger2010early}) for preliminary image processing to extract features with a stereo camera. The features are edges, contours, textures, and surfaces. The robot tries to grasp different objects and associates ECV's features to successful grasps. A limitation of this work is that ECV needs textured or complex objects to work properly. 

Those works are conceptually similar to ours: a robotic system explores an environment (here an object)  with an action (here grasping)  and learns to associate local visual features to successful actions. However, they assume that the system is already able to extract objects from a scene and focus on it to learn grasping. In our work, the robotic system has no notion of objects. The whole environment is considered, in order to learn relevant areas for different affordances. From these areas, object candidates could be extracted as a base for the above-mentioned methods. Thus, these works correspond to a later developmental step with respect to ours.

Ugur et al. \cite{Ugur2007} proposed a method for learning "traversability" affordance with a wheeled mobile robot which explores a simulated environment. The robot tries to go through different obstacles: laying down cylinders, upright cylinders, rectangular boxes, and spheres. The laying down cylinders and spheres are traversable while boxes and upright cylinders are not. The robot is equipped with a 3D sensor and collects data after each action labeled with the success of going through the objects. The sample data are extracted thanks to a simulated RGB-D camera. Then, an online SVM (\cite{bordes2005fast}) is trained based on the collected data. The resulting model predicts the "traversability" of objects based on local features. To drive the exploration, an uncertainty measure is computed based on the soft margin of the model decision hyperplane. Finally, they tested their method on a navigation problem, on real robots and in a realistic environment. They demonstrate, by using the model learned in simulation, that the robot is able to navigate through a room full of boxes, spherical objects and cylindrical objects like trash bins without colliding with non-traversable objects.

Kim and Sukhatme \cite{kim2015interactive}, with a similar idea, seek to learn pushable objects in a simulated environment using a PR2 with an RGB-D camera. The objects are blocks the size of the robot. They are either pushable in one or two directions, or not pushable. The PR2 uses its two arms to try to push the blocks. The learning process relies on a logistic regression classifier and a Markov random field is used to smooth spatially the predictions. The robot explores then the environment and collects data by trying to push the blocks. The outcome of the framework is what they called an affordance map indicating the probability of pushability of a block. When in the work of U\v{g}ur et al. \cite{Ugur2007} the learning is made on continuous space, in the work of Kim and Sukathme \cite{kim2015interactive} the environment is discretized in a grid with the cells of the size of a block, thus, the learning space is discrete. Finally, they use an exploration strategy based on uncertainty reduction to select the next block to interact with.

In a more developmental perspective, Paletta et al. \cite{paletta2007perception} proposed a framework to learn composite affordances by starting from low level affordances. Their approach is split into 3 steps: first, the robot explores its environment with a reactive behavior, like a grasp reflex, and collects visual data consisting of SIFT. Then, in a second step, basic affordances are learned with simple actions such as pushing or gripping. Finally, in the third step, the robot learns composite affordances based on a combination of the basic action used in the previous step. For instance, this combination of actions allows the robot to achieve stacking.
They validate their framework with a mobile robot equipped with a stereo camera and a magnetized end-effector. In a real environment the robot tries to learn to identify objects that are liftable with its magnetized end-effector.

These works \cite{Ugur2007,kim2015interactive,paletta2007perception} are close to the work presented in this paper. They gather in a single study affordance learning, online learning, exploration process, and interactive perception. The affordance map of Kim and Sukhatme \cite{kim2015interactive} is close to our relevance map by the way they both segment interesting elements for the agent, but exploration and learning were conducted in simulation only, in simple environments and setups, and only one affordance was learnt. The study proposed by Paletta et al. \cite{paletta2007perception} can learn several affordances in simulation, but it was tested in reality with only one action. The approach proposed in this article is based on similar principles but it allows the system to learn relevance maps relative to several affordances in more complex and realistic environments, in real world-experiments.

\section{Method}\label{sec:method}

The goal of this work is, for a robot, to learn which part of an environment affords a given effect to a specific action through an autonomous exploration.
 The robot is interacting with the environment thanks to an action primitive in order to collect data. The method is tested with three affordances: pushable objects, activable push-buttons and liftable objects. These affordances are respectively linked to a push primitive, a push-button primitive and a lift primitive.  
\begin{figure}[!h]
\centering
\includegraphics[width=\linewidth]{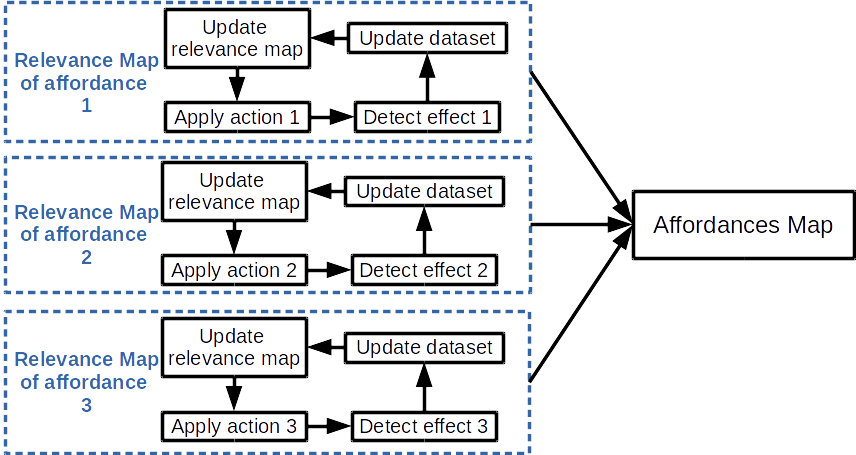}
\caption{Overview of the general approach to build an affordances map}
\label{fig:schema_gen}
\end{figure}

The general approach, summarized in figure \ref{fig:schema_gen}, is to separately build the relevance map relative to each considered affordance. Each relevance map is built by collecting data thanks to the interactions of the robot and then by training online a classifier on the data. The classifier is used to build the relevance map by attributing weights to segments extracted from the current scene (see sections \ref{sec:cmm} and \ref{sec:sv}). Finally each relevance map is merged in one affordances map (see section \ref{sec:am}).    

A formalization  of affordances is proposed in section \ref{sec:aff_form}. Then, in section \ref{sec:wf}, the workflow to build a relevance map is explained. Finally, the classifier is described in section \ref{sec:cmm}.

\subsection{Affordances Formalisation}\label{sec:aff_form}

In this study, an affordance is a relation $\phi$ between an action $a$ and an effect $e$. This relation is formalized as a conditional probability of an effect $e$ to occur after the application of an action $a$ on an element with a visual feature $X$ (see equation \ref{eq:aff_proba}). Thus, $\phi$ is a function parameterized by $a$ and $e$ which takes as input $X$ and gives as output a value between 0 and 1. This value represents the probability of existence of the affordance $(a,e)$ on $X$.  

\begin{equation}\label{eq:aff_proba}
\phi_{(a,e)}(X) =  P(\Delta = (a,e) | X) 
\end{equation}  
%

With this formalization, we define \textit{composite affordances} as a composition of one or several affordances. 
\begin{equation}\label{eq:aff_comp}
\begin{split}
P(\Delta_1 | X) & = P(\Delta_1 | X, \Delta_0) P(\Delta_0 | X) \\
\phi_{(a_1,e_1)}(X) & = P(\Delta_1 = (a_1,e_1) | X, \Delta_0 = (a_0,e_0)) \phi_{(a_0,e_0)}(X)
\end{split}
\end{equation}
Equation \ref{eq:aff_comp} presents the formal representation of a composite affordance which links an action $a_1$ and an effect $e_1$ to an action $a_0$ and an effect $e_0$. The relation is defined thanks to the Bayes' rule. This proposition means that if the feature $X$ affords the action $a_1$ by producing the effect $e_1$ then it affords the action $a_0$ by producing the effect $e_0$ too. In other words, to exist, the affordance $(a_1,e_1)$ needs the existence of the affordance $(a_0,e_0)$. In the following text, we say that the probability of $X$ to afford $a_1$ by producing the effect $e_1$ is \textit{filtered} by the probability of $X$ to afford $a_0$ by producing the effect $e_0$.

Equation \ref{eq:aff_comp_multi} presents the general case of a composite affordance as a composition of several other affordances. For this equation to be true, all the component affordances must be independent from each other.

\begin{equation}\label{eq:aff_comp_multi}
\begin{split}
\phi_{(a,e)}(X) & =  \\
P(\Delta  = (a,e)| X, \bigcap^{n}_{i=0} \Delta_i & = (a_i,e_i)) \prod^{n}_{i=0} \phi_{(a_i,e_i)}(X)
\end{split}
\end{equation}

For instance, in this article, the probability of something to be liftable is filtered by the probability of something to be pushable. Because we assume that something liftable is also pushable, thus the liftable affordance requires the pushable affordance.

\subsection{Workflow to build a relevance map}\label{sec:wf}

\subsubsection{Overview}

Our method aims at building an affordances map through an autonomous exploration driven by a robot equipped with two arms. The affordances map is the combination of several relevance maps. Each of them is relative to a specific affordance. To build one relevance map, the robot explores the environment which is unknown, with a specific action primitive. The system detects a possible effect thanks to an effect detector specific to the action primitive. Thanks to the interaction and the effect detector, labeled samples are collected. They are labeled with a value of 1 if the interaction produced an effect, with a value of 0 otherwise. The classifier is used to build the relevance map. The classifier is trained online to allow the relevance map to drive the exploration to be efficient. The visual system of the robot is an RGB-D camera (Microsoft Kinect v2) which generates 3D pointclouds. 

\begin{figure}[!h]
\centering
\includegraphics[width=\linewidth]{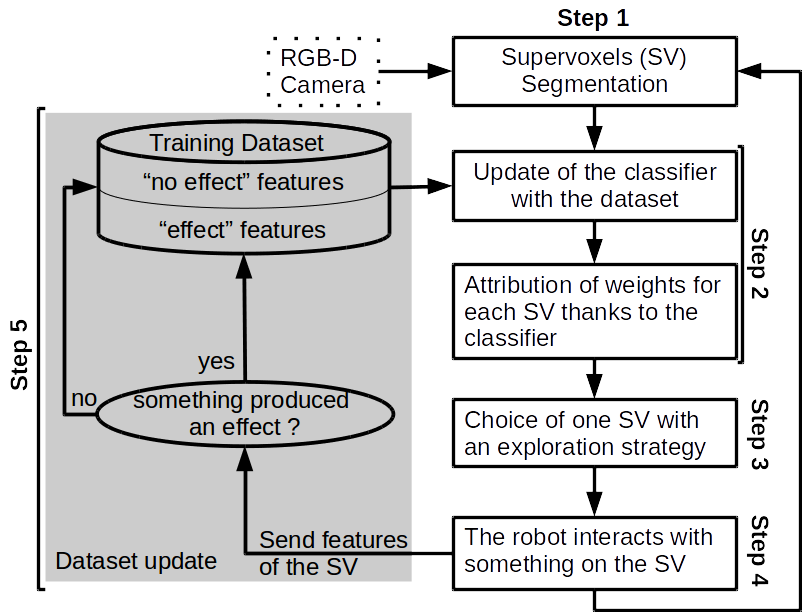}
\caption{Overview of the workflow followed during an exploration to build a relevance map}
\label{fig:schema_wf}
\end{figure}

The exploration is sequential, the robot interacts with the environment, observes the effect, updates its perception and starts again. During the interaction, the system does not update its perception.
The workflow of one iteration (shown in figure \ref{fig:schema_wf}) follows 5 steps :
\begin{itemize}
\item \textbf{Step 1:}  An oversegmentation of the 3D pointcloud into \textit{supervoxels} using Voxel Cloud Connectivity Segmentation (VCCS) method is done on the current scene. Visual features are extracted from each supervoxels. Supervoxel segmentation is described in section \ref{sec:sv} and the visual feature extraction method is explained in section \ref{sec:feat}.
\item \textbf{Step 2:}  The classifier is updated with the training dataset extended with a new labeled sample. Then, the classifier weights are attributed to each supervoxel. The outcome is the relevance map of the current scene. This step is explained in section \ref{sec:rm}.
\item \textbf{Step 3:}  The next supervoxel to interact with is chosen as a target for the action primitive. Section \ref{sec:choice} explains how the target is chosen.
\item \textbf{Step 4:}  An action primitive is applied on the center of the chosen supervoxel. Each action primitive is explained in section \ref{sec:ap_ed}.
\item \textbf{Step 5:}  To check if an effect is produced by the action primitive, an effect detector is applied. The visual feature of the selected supervoxel is added to the training dataset with a label indicating if there was an effect. The different effect detectors are described in section \ref{sec:ap_ed}.
\end{itemize}

\subsubsection{Supervoxels}\label{sec:sv}

The relevance map relies on supervoxels segmentation. Supervoxels were introduced by J. Papon \cite{Papon2013} with his voxel cloud connectivity segmentation (VCCS) method. A supervoxel is similar to a superpixel like in SLIC  \cite{Achanta2010} or turbopixel \cite{levinshtein2009turbopixels} methods except that it integrates depth information. Contrary to superpixel segmentation, VCCS works directly on a 3D pointcloud. A supervoxel is a cluster of voxels. A voxel is the smallest unit in a 3D image. In a pointcloud, a voxel is a point. The use of depth information allows the supervoxels to respect the boundary of objects which is a significant enhancement compared with superpixels. So, the information extracted from a supervoxel is more likely to be relative to a single component of the environment. Thus, this information is more consistent. 

VCCS method workflow is the following : voxel seeds are evenly distributed on the pointcloud, then with local nearest neighbor, regions grow from these seeds by adding voxels. The neighborhood is defined thanks to a radius named seed radius ($R_{seed}$). This hyperparameter controls the size of the supervoxels. The local nearest neighbor uses a distance (see equation \ref{eq:vccsdist}) composed of CIELab\footnote{CIELab is a colorimetry international standard from the International Commission on Illumination (CIE) of 1978} color distance, spatial distance, and shape distance computed thanks to the fast point feature histogram (FPFH) \cite{rusu2009fast} algorithm. 
	
	\begin{equation}\label{eq:vccsdist}
		D = \sqrt{\frac{\lambda D_c^2}{m^2} + \frac{\mu D_s^2}{3 R_{seed}^2} + \epsilon D^2_{f}}
	\end{equation}

As shown in equation \ref{eq:vccsdist}, three weights $\lambda$, $\mu$, and $\epsilon$ control the importance of each distance. Therefore, VCCS algorithm has four important hyperparameters. $R_{seed}$ controls the size of the supervoxels and ($\lambda$, $\mu$, $\epsilon$) control their shapes. Only the size of the supervoxel is critical because if an object is smaller than a supervoxel then the information extracted from it will not be consistent. While, for the three other parameters, they can be tuned to have meaningful supervoxels for a large range of environments. 

A major drawback of VCCS is the inconsistency of the segmentation over a video stream. When extracted on a video stream, the segmentation is different for each frame even if the scene is static. This due to the noise of the depth image. 

In this work, supervoxels are used as the smallest visual unit for image processing as well as for the action primitives targets. The version of VCCS implemented in the PointCloud Library is used \cite{Rusu_ICRA2011_PCL}.

In this implementation, the algorithm gives as output a centroid point for each supervoxels which is at the average position, has the average color and normal of the points in the supervoxel. Also, an adjacency map is provided which represents a graph of euclidean proximity of each supervoxel. Therefore, to find the neighbors of a supervoxel, going through the adjacency map is enough.

\subsubsection{Features Extraction}\label{sec:feat}

The visual features extracted from the supervoxels and used to train the classifier are the concatenation of color histrograms with the CIELab encoding and a geometric descriptor based on FPFH. 

For each channel of the CIELab color, a five-bin histogram is extracted. Then, they are concatenated into one vector of 15 entries. 

Fast point feature histogram (FPFH) proposed by R.B. Rusu et al. \cite{rusu2009fast} is a widely used geometrical descriptor. It is appreciated for its high discriminative capacity. In the present method, FPFH is extracted on the central point of the pointcloud including the targeted supervoxel and its neighbors. The radius of neighborhood to compute FPFH is set to the size of a supervoxel, thus the central point FPFH takes into account the whole considered pointcloud. The central point is the centroid of the targeted supervoxel. This feature has 33 dimensions.

\begin{figure}[h]
\centering
\includegraphics[width=.9\linewidth]{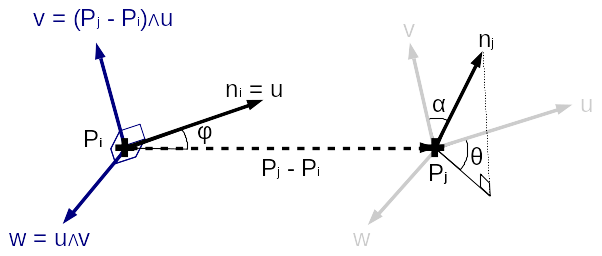}
\caption{Schema of how the orthogonal frame $(u,v,w)$ is defined on which the computation of SPFH is based. Figure reproduced from \cite{rusu2009fast}.} 
\label{fig:spfh}
\end{figure}

FPFH is a modification of PFH (point feature histogram) from the same authors to be computationally faster. To compute an FPFH an a target point, simplified PFHs (SPFH) are computed on the target point, on its neighbors and on the neighbors of its neighbors. The neighborhood is defined according to a radius which is an hyperparameter of the algorithm. Therefore, FPFH includes information of points within two times the neighborhood radius. An SPFH is a histogram on three angles $\alpha$, $\phi$, and $\theta$. Equations system \ref{eq:spfh} shows how these angles are computed. 
\begin{equation}\label{eq:spfh}
\begin{split}
\alpha & = v*n_j \\
\phi & = u*\frac{p_j-p_i}{\Vert p_j - p_i\Vert} \\
\theta & = arctan(w*n_j,u*n_j) 
\end{split}
\end{equation}
Where $(.*.)$ is the scalar product, $(u,v,w)$ is an orthogonal frame defined in equation \ref{eq:spfhframe} and shown in figure \ref{fig:spfh}, $n_i$ and $n_j$ are the normals to the surface at the points $p_i$ and $p_j$.

\begin{equation}\label{eq:spfhframe}
\begin{split}
u & = n_s \\
v & = u \wedge \frac{p_t-p_s}{\Vert p_t - p_s \Vert} \\
w & = u \wedge v
\end{split}
\end{equation}	
Where $(. \wedge .)$ is the vectorial product.

\subsubsection{Building a Relevance Map}\label{sec:rm}

Thanks to the online trained classifier, the supervoxels are weighted with values between 0 and 1. A weight represents the relevance of a supervoxel, i.e. the probability of a supervoxel to be part of a component which will produce an effect after the application of a certain action. Thus, a relevance map is a set of weighted supervoxels. The classifier named collaborative mixture models is described in section \ref{sec:cmm}.

\subsubsection{Choice of the Next Area to Explore}\label{sec:choice}

From the predictions of the classifier, a choice distribution map is computed. The choice distribution map is also a set of weighted supervoxels, but a weight represents the probability for the supervoxel to be chosen by the system as the next target of the interaction. A weight is the combination of the \textit{uncertainty} and the \textit{confidence} of the classifier and the \textit{diversity} of the dataset. The higher the uncertainty and the lower the confidence and the diversity, the higher is the probability for a supervoxel to be chosen. This step is described in section \ref{sec:cdm}.

\subsubsection{Action Primitives and Effect Detectors}\label{sec:ap_ed}

\paragraph*{Pushable Objects}

The pushable affordance is associated to a push primitive and a change detector as effect detector. The push primitive is going through three steps. First, the end-effector is going in an approach pose near and oriented towards the target. Then, the end-effector follows a straight line towards the target until going through it. Thus, if a pushable object is on the target, it will be pushed. Finally, a reverse motion is applied in which the arm goes back to its home position. For each interaction, the left or the right arm is randomly chosen. If no valid plan is found with the chosen arm then planning is tried with the second arm. 

The planning is done within a framework called MoveIt \cite{sucan2012ompl,moveit}. This framework provides planning algorithms with obstacle avoidance. Obstacle avoidance is used during the approach motion to prevent any involuntary disturbance in the scene. 

\begin{figure}[!h]
\centering
\includegraphics[width=.9\linewidth]{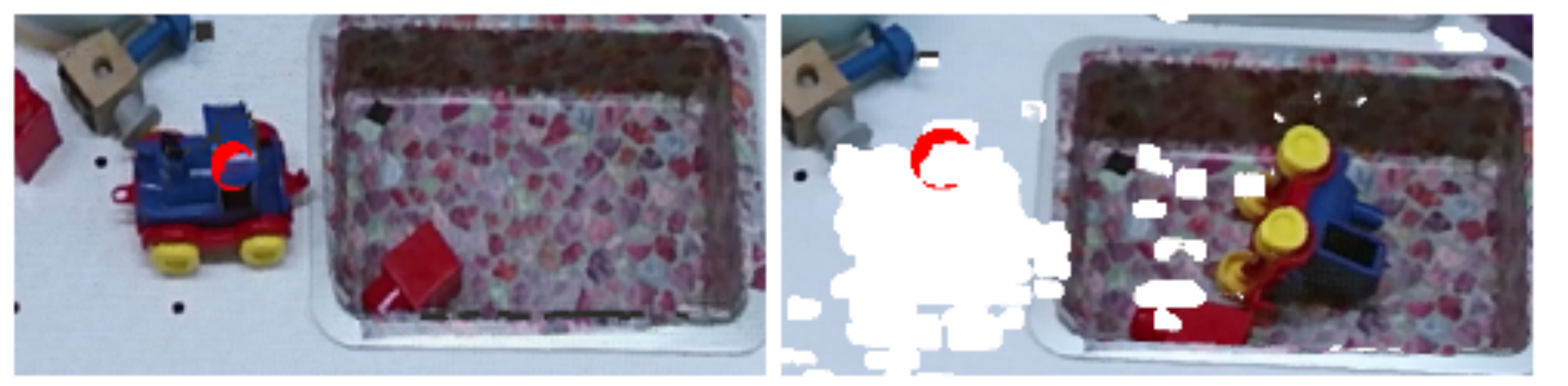}
\caption{Visualisation of the change detector. The right picture represents a part of a scene before a push and the left picture after a push. The red dot on both pictures represents the target of the push primitive which is here the upper part of the blue toy. This target corresponds to the center of a supervoxel. The white areas represent the parts detected as different between both images.}
\label{fig:chdet}
\end{figure} 

The effect detector is a simple change detector of the scene.
As shown in the figure \ref{fig:chdet}, a difference point cloud is computed (in white in the figure \ref{fig:chdet}) by substracting the pointcloud before (the left picture of the figure \ref{fig:chdet}) and the one after the interaction (the right picture of the figure \ref{fig:chdet}). Then, if the points of the targeted supervoxel (its center is represented by a red dot in the figure \ref{fig:chdet}) is part of the difference pointcloud then a change has occured.

\paragraph*{Activable Push-Button}
This affordance is associated with push-buttons which activate a signal displayed on a screen visible to the robot. The action primitive is similar to the push primitive except for the orientation which is only vertical or horizontal in the robot frame (the push primitive used to learn the pushable affordance has a continuous range of orientations). The effect detector is a recognition system which allows the robot to see if a button is pushed. The state of the buttons is displayed on a screen like in the pictures of figure \ref{fig:modules}. The state is perceived by the robot thanks to a visual recognition system implemented with OpenCV. This system is specific to the interface.

\begin{figure}[!h]
\centering
\subfloat[No button pushed]{\label{fig:button_not_pushed}
\includegraphics[width=.3\linewidth]{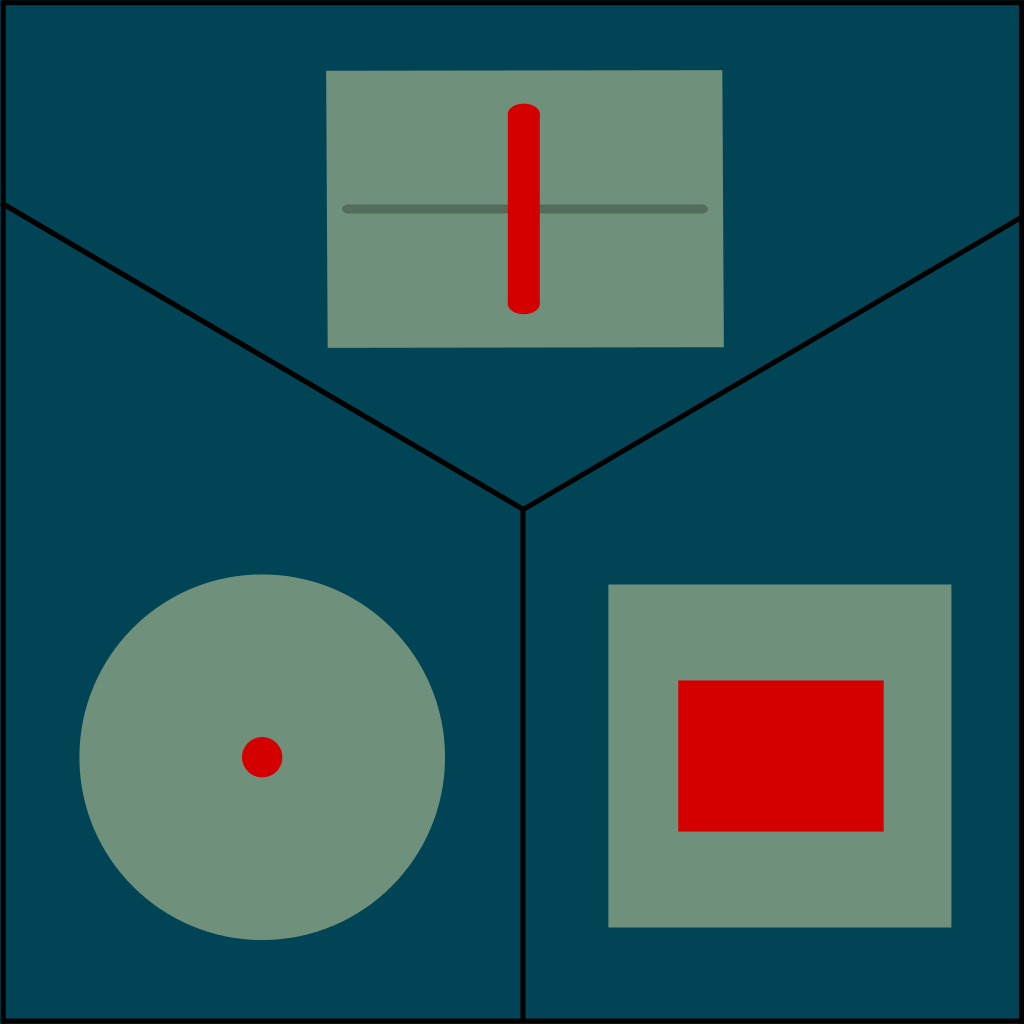}
}
\subfloat[At least button is pushed]{\label{fig:button_pushed}
\includegraphics[width=.3\linewidth]{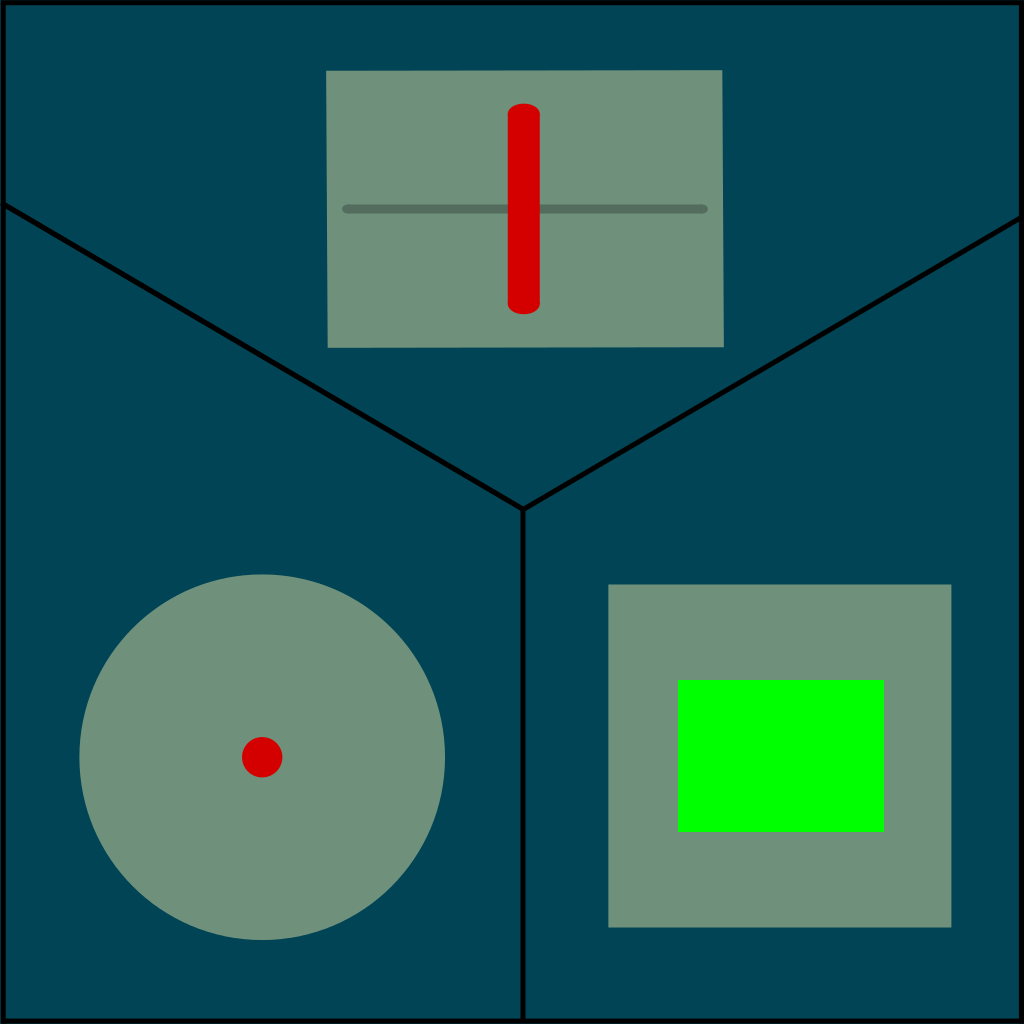}
}
\caption{Interface which displays on a screen the state of different interactive modules. For the present study, only the right bottom part is used. It displays the buttons state. The rectangle is red if no button is pushed and it becomes green if at least one button is pushed.}
\label{fig:modules}
\end{figure}

\paragraph*{Liftable Objects}

Among the pushable parts in the environment the robot will try to learn liftable parts. It is assumed that liftable parts are first pushable, thus, liftable affordance is a composite affordance, composed by the pushable affordance. The probability to afford the lift primitive is filtered by an already learned probability to afford the push primitive. Therefore, the exploration is biased by a relevance map of pushable affordance. 

For this affordance, the robot uses a lift primitive that consists in going above the target, rotating the wrist of its gripper in a certain orientation, then going down and closing the gripper before finally going up again and letting the lifted "thing" fall by reopening the gripper. Like for the push primitive during the approach motion to go above the target, the obstacles are avoided.

To detect if something is lifted, the opening of the gripper is checked before reopening the gripper. If it is not fully closed, the target will be considered as lifted.

In this primitive the gripper is fixed in the vertical orientation, thus, only liftable objects laying on a horizontal plane are considered here. The approach can be extended to any liftable object with an appropriate lift primitive.  

\subsection{Building the affordances map}\label{sec:am}

The affordances map is a combination of several relevance maps. In this way, each supervoxel has a set of weights assigned corresponding to each relevance. All the weights under a certain threshold are reduced to zero. The affordances map is represented by assigning a color to each affordance and no color for supervoxels with weights all equal to zero.

\subsection{Collaborative Mixture Models}\label{sec:cmm}

This section presents the classifier used in this article : the Collaborative Mixture Models (CMMs). This classifier was introduced in our previous work \cite{legoff2019boostrapping}. 

\subsubsection{Definition of the classifier}

The conditional probability which formalized an affordance is the output of CMMs. CMMs are used to classify samples between two classes $(a,e)$ and $(a,\overline{e})$. The first one is the class of \textit{effect occurance} and the second one is the class of \textit{absence of effect} after the application of action $a$. Equation \ref{eq:class_proba} defines the probability of a features $X$ to be part of the class $(a,e)$.

\begin{equation}\label{eq:class_proba}
P(\Delta = (a,e) | W, \Theta, X) = \frac{1 + \Gamma(W_e,\Theta_e,X)}{2 +  \Gamma(W_e,\Theta_e,X) + \Gamma(W_{\overline{e}},\Theta_{\overline{e}},X)}
\end{equation}
Where $\Gamma$ is a Gaussian mixture model (GMM), $W_e$ are the weights associated to the GMM of class $(a,e)$, $\Theta_e$ are the parameters of the multivariate normal distributions of the GMM associated to $(a,e)$, $W = W_e \cup W_{\overline{e}}$, and $\Theta = \Theta_e \cup \Theta_{\overline{e}}$. 
1 is added to the numerator and 2 is added to the denominator to obtain a default probability of $\frac{1}{2}$ if both mixtures are empty.

The parameters of CMMs are the following :
\begin{itemize}
\item $K_E$: number of components of the mixture models encoding the class $\Delta = (a,E)$ with $E \in \{e,\overline{e}\}$. This number is estimated during the training.
\item $S = \{s_i,\Delta_i\}_{i<I}$: database of samples and their corresponding label constituted during the exploration. 
\item $\Theta_E = \{\mu_k,\Sigma_k\}_{k<K_E}$: parameters of the multivariate normal distribution of model associated to class $\Delta$ with mean $\mu_k$  and covariance matrix $\Sigma_k$. They are estimated thanks to their sample estimator.
\item $W_E = \{w_k\}_{k<K_E}$: weights of the mixture model associated with class $\Delta$. These parameters are computed thanks to equation \ref{eq:weights}.
\item $\Delta \in \{(a,e),(a,\overline{e})\}$: class to be predicted by the classifier. 
\end{itemize} 

Therefore, each class is modeled by a GMM. And each GMM is composed of several multivariate normal distributions. A distribution models a component. The means and the covariances of the distributions are computed thanks to their samples estimators. 

A component is a cluster of samples of the feature space modeled by a multivariate normal distribution. Naturally, all the samples of a component have the same label. So formally, we write $C_k(X) = (w_k,G(\mu_k,\Sigma_k,X)),S_k,\Delta)$ a component $k$, where $S_k$ is the dataset used to estimate $\mu_k$, $\Sigma_k$, and $w_k$, and $G$ is a Gaussian function. 
The probability of a sample with feature $X$ to belong to a component $k$ is given by the equation \ref{eq:comp_est}. 

\begin{equation}\label{eq:comp_est}
P(K_E=k|X,\Theta,\Delta) = \frac{w_k*G(\mu_k,\Sigma_k,X)}{\sum^{K_E-1}_{i=0}{w_i*G(\mu_i,\Sigma_i,X)}}
\end{equation}

Let $M_E = \{C_k\}_{k<K_E}$ be the set of components of class $\Delta$.

The weights of a GMM are computed following the equation \ref{eq:weights}.
\begin{equation}\label{eq:weights}
w_k = \frac{\vert C_k \vert}{\sum_i^{K_E} \vert C_i \vert} 
\end{equation}

\subsubsection{Training Algorithm}

CMMs is trained online and in a supervised way. When a new sample arrives, the algorithm follows these three steps:
\begin{itemize}
\item[1.] If no component of the class of the new sample have been created yet, then a new component is created with as center the feature of the sample and a covariance equal to an identity matrix multiplied by a constant. If at least one component exists then the sample is added to the closest component and the parameters of this component are updated.
\item[2.] A \textit{split} operation is applied on the updated component, if it is not split then the \textit{merge} operation is applied.
\item[3.] A \textit{split} operation is applied on a random component of each class, if they are not split then the \textit{merge} operation is applied.
\end{itemize}

\begin{figure}[!h]
\centering
\subfloat[Condition for a split: if two components of different classes are overlapping.]{\label{fig:split_illu}
\includegraphics[width=.8\linewidth]{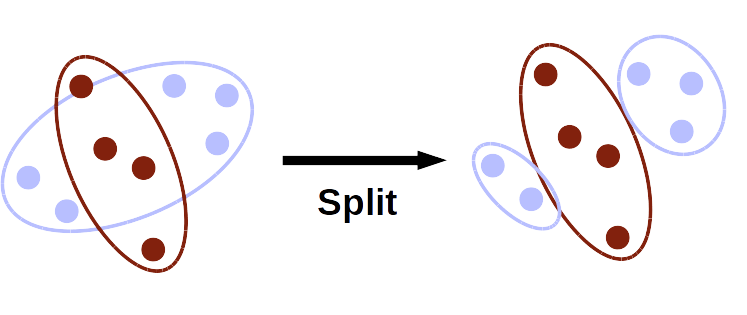}
} \\
\subfloat[Condition for a merge: if two components of the same class are overlapping]{\label{fig:merge_illu}
\includegraphics[width=.8\linewidth]{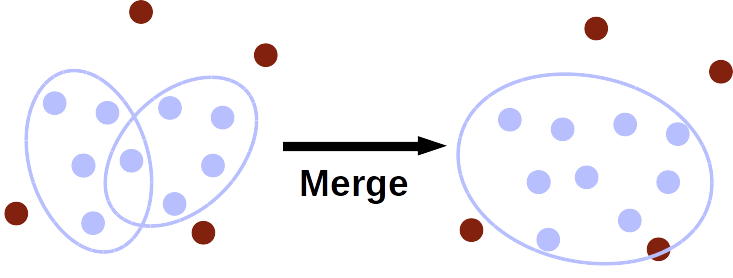}
}
\caption{Illustration of the conditions of split and merge application.}
\label{fig:modules}
\end{figure}

The split operation increases the number of components to adapt the distribution to the dataset and the merge operation reduces the number of components to avoid overfitting and reduce computational time. Split and merge operations are described respectively in algorithm \ref{algo:split} and \ref{algo:merge}.
A split operation occurs if two components of different classes are overlapping, as illustrated in figure \ref{fig:split_illu}. A merge operation occurs if two components of the same class are overlapping as illustrated in figure \ref{fig:merge_illu}. The number of components is limited to a maximum number. If this number is reached only the merge operation is applied.

To decide if two components are overlapping, hyperellipsoids of tolerance of the multivariate normal distributions are estimated. Then, intersection between the ellipsoids is checked. The intersection condition, shown in equation \ref{eq:intercond}, is a simplification of a complete intersection condition.

\begin{equation}\label{eq:intercond}
(\rho - \mu)^{T}\Sigma^{-1}(\rho - \mu) <= \frac{(n-1)p}{n-p}\frac{n+1}{n}F_{1-\alpha}(p,n-p) 
\end{equation}
Where $\mu$ and $\Sigma$ are the mean and covariance of the candidate component to be split or merged, $\rho$ the mean of another component, $n$ the number of samples present in the candidate component, $p$ the dimensionality of the feature space (48 in our case) and $F_{1-\alpha}$ the quantile function of the Fisher distribution. The argument of this function must be strictly above 0, therefore $n$ must be strictly greater than $p$. This constraint makes the number of samples at least equal to 48 in the candidate component necessary. 

$\alpha$ determines the size of the ellipsoid : the ellipsoid encloses all the samples with a probability above $1 - \alpha$. Thus, the smaller $\alpha$, the bigger is the ellipsoid. $\alpha$ controls the intersection condition sensibility. If $\alpha$ is equal to 1, no intersection is considered and so no split and merge operations are applied.


\begin{algorithm}[h]
\caption{SPLIT algorithm}\label{algo:split}
\begin{algorithmic}[1]
\Procedure {SPLIT}{$C$,E,$M_e$,$M_{\overline{e}}$}
 \If{$|M_E|<K_{max}$} \Comment{\small{If the number of components of class $\Delta$ is above $K_{max}$}}
  \State \Return $M = M_e\cup M_{\overline{e}}$ \Comment{Then abandon the split} 
 \EndIf 
 \State $C' \leftarrow closest\_component(C) \in M\setminus\{M_E\}$ \Comment{Search the closest component from $C$ from a class $\neq \Delta$}
  \If{$C' \cap C \neq \emptyset$} \Comment{If component C intersect with C'}
   \State $C_1, C_2$ = $split(C)$ 
   \State $M_E \leftarrow (M_E \setminus \{C\}) \cup \{C_1,C_2\}$ 
 \EndIf
\State \Return $M = M_e\cup M_{\overline{e}}$
\EndProcedure
\end{algorithmic}
\end{algorithm}

\begin{figure}[h]
\centering
\subfloat[Step 1]{\label{fig:split1}
\includegraphics[height=60mm,width=.3\linewidth,keepaspectratio]{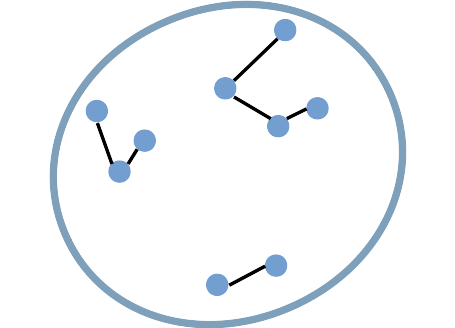}
} 
\subfloat[Step 2]{\label{fig:split2}
\includegraphics[height=60mm,width=.3\linewidth,keepaspectratio]{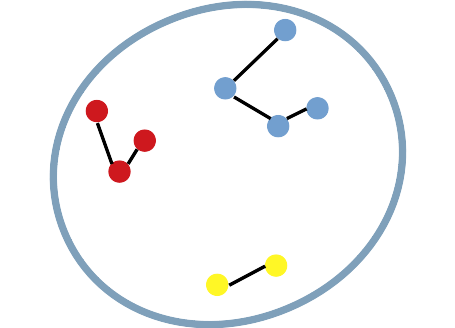}
} 
\subfloat[Step 3]{\label{fig:split3}
\includegraphics[height=60mm,width=.3\linewidth,keepaspectratio]{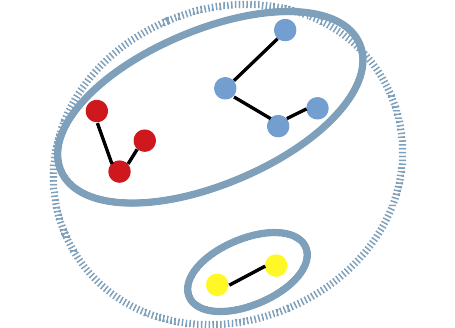}
} 
\caption{Illustration of how the samples are shared between two new components during a split.}
\label{fig:split_schema}
\end{figure}

To share the samples of the split component between two new components the following algorithm is used (illustrated in figure \ref{fig:split_schema}): 
\begin{itemize}
\item[1.] Build a graph of minimal distances between the features of the samples of the components;
\item[2.] Group the samples per connected sub-graph;
\begin{itemize}
\item If there is only one group then cancel the split;
\item If there are two groups then go to step 3; 
\item If there are more than two groups then merge the closest group until having two groups and go to step 3.
\end{itemize}
\item[3.] Create two new components from the samples of the two groups.
\end{itemize}

\begin{algorithm}[h]
\caption{MERGE algorithm}\label{algo:merge}
\begin{algorithmic}[1]
\Procedure {MERGE}{$C$,E,$M_e$,$M_{\overline{e}}$}
 \State $C' \leftarrow closest\_component(C) \in M_E$ \Comment{Search the closest component from $C$ in $M_E$}
  \If{$C \cap C' \neq \emptyset$} \Comment{If component C intersect with C'}
   \State $\tilde{C} \leftarrow C \cup C'$ 
   \State $M_l \leftarrow (M_l \setminus \ {C,C'}) \cup \tilde{C}$
  \EndIf
\State \Return $M = M_e\cup M_{\overline{e}}$ 
\EndProcedure
\end{algorithmic}
\end{algorithm}  
 
CMMs have two hyperparameters :
\begin{itemize}
\item $\alpha \in [0,1]$ which controls the interaction condition sensibility. It is fix at 0.6 for all the experiments. 
\item $K_{max} \in N$ which is a maximum number of components in each class. It appears in the split operations algorithm (\ref{algo:split}). It is fixed at 4 for all the experiments. 
\end{itemize}

\subsubsection{Choice Distribution Map}\label{sec:cdm}

From the classifier, two metrics are computed to drive the exploration of the robotic system : the \textit{uncertainty} and \textit{confidence} of classification. They are combined to output a probability of choice of a feature $X_i$ of the $i^{th}$ supervoxel extracted on a pointcloud as shown in equation \ref{eq:samplproc}. 

\begin{equation}\label{eq:samplproc}
P_c(X_i) = u(X_i)*(1-c(X_i))
\end{equation}  
Where $u(.)$ is the uncertainty and $c(.)$ is the confidence. 

\begin{figure}[h!]
\centering
\includegraphics[width=.8\linewidth]{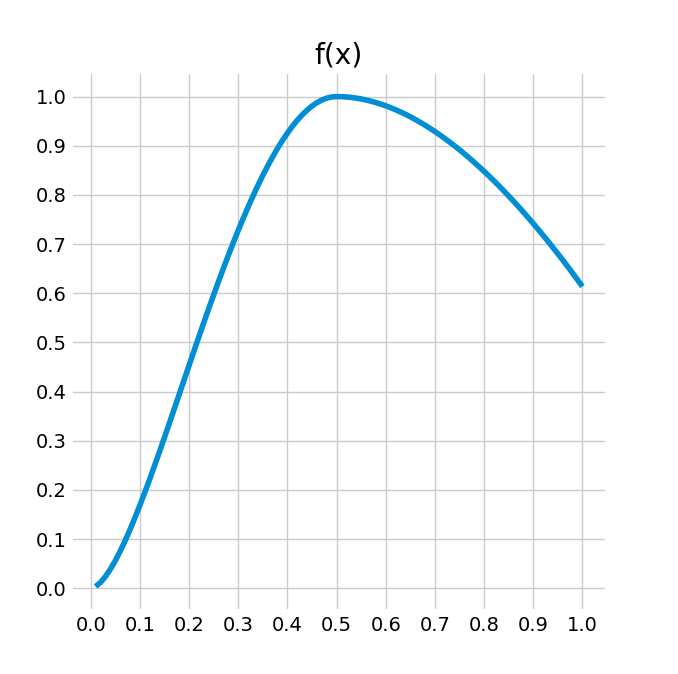}
\caption{Function used for uncertainty estimation. This function gives a higher probability of choice to uncertain classification, but also to certain classification to the chosen class, i.e the one with fewest samples.}
\label{fig:xlogx}
\end{figure}

\paragraph*{Uncertainty} As CMMs is a probabilistic classifier, its output can give directly an uncertainty measure. The output of CMMs (see equation \ref{eq:comp_est}) is a probability of membership of a sample in a class. The closer this probability from 0.5, the more uncertain the classification is. Uncertainty of classification is computed thanks  to the equations \ref{eq:u} and \ref{eq:xlogx} 

\begin{equation}\label{eq:u}
u(X_i) = 
\begin{cases}
f(p) & \vert S_e \vert <= \vert S_{\overline{e}} \vert \\
f(1-p) & \vert S_e \vert > \vert S_{\overline{e}} \vert
\end{cases}
\end{equation}
where $p = P(\Delta = (a,e) | W, \Theta, X)$ and $f$ is the following function:

\begin{equation}\label{eq:xlogx}
f(x) =
\begin{cases}
 -2x(log(2x)-1) & x >= 0.5 \\
 -4x^2(log(4x^2)-1) & x < 0.5
\end{cases}
\end{equation}
The function $f(.)$ is plotted in figure \ref{fig:xlogx}.

The uncertainty computed this way drives the exploration to collect samples with features from uncertain area in the feature space. Also, the exploration gives priority to class with less samples collected over other classes (see figure \ref{eq:u}). This pushes the system to build a balanced dataset between the classes. 

\paragraph*{Confidence} The classification of CMMs relies on a mapping of the feature space of normal distributions. The border of these distributions can give an insight on the least dense areas in the feature space. The confidence of the classifier for a sample is its probability of membership in its closest component. This probability is computed thanks to equation \ref{eq:comp_est}. By choosing areas with the lowest confidence, the exploration gives a focus to areas in which the system has less information. Therefore, this metric could be interpreted as an approximation of entropy.

\section{Experiments}\label{sec:exp}

\subsection{Protocol}

\begin{figure}[!h]
\centering
\includegraphics[width=.8\linewidth]{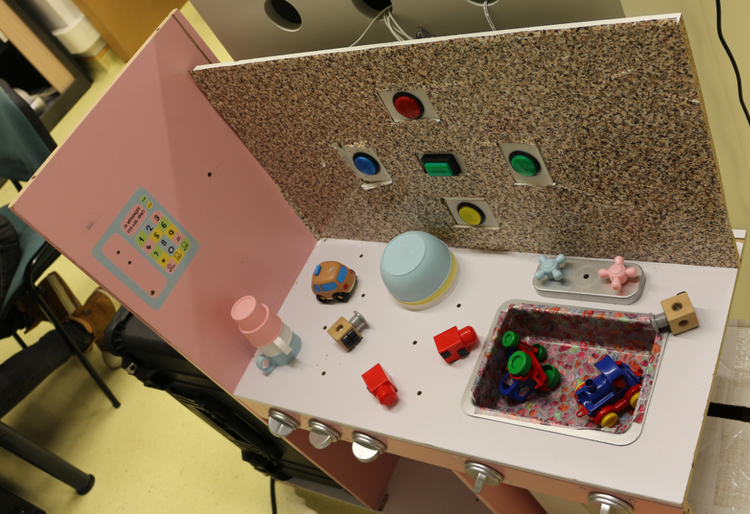}
\caption{The setup used for all the experiments. The setup is a toy kitchen with 5 interactive push buttons integrated into a vertical plane.}
\label{fig:aff_env}
\end{figure}

For each of the three affordances, 4 experiments have been conducted. 
An initialization step has been added in the experiments of liftable and activable push-button in which the system is forced to gather at least 10 samples of each class. With a uniform random sampling, the chance to gather positive samples in these experiments is very low, thus at the beginning of the experiment, the robot collects only negative samples. This initial step allows the system to start from a balanced dataset. Adding this step was not useful for the experiments with the push affordance as the probability to gather positive samples is higher.  

\begin{figure}[!h]
\centering
\subfloat[\tiny{Toy locomotive : \textbf{pushable}, \textbf{liftable}}]{\label{fig:loco}
\includegraphics[width=.2\linewidth]{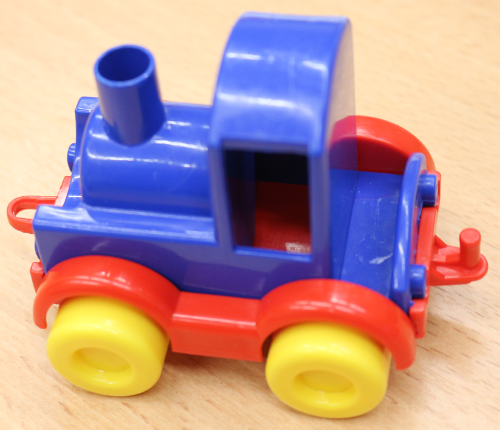}
} 
\subfloat[\tiny{Toy locomotive : \textbf{pushable}, \textbf{liftable}}]{\label{fig:loco2}
\includegraphics[width=.2\linewidth]{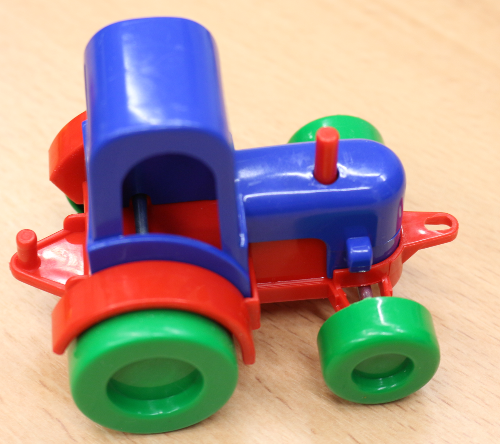}
} 
\subfloat[\tiny{Pile of bowls : \textbf{pushable}}]{\label{fig:bowl}
\includegraphics[width=.2\linewidth]{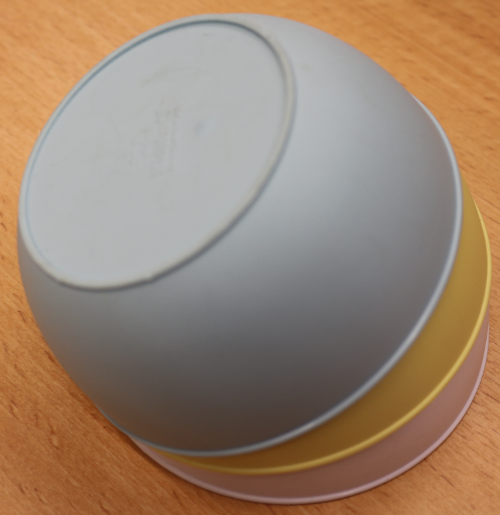}
} 
\subfloat[\tiny{Pile of mugs : \textbf{pushable}}]{\label{fig:mug}
\includegraphics[width=.15\linewidth]{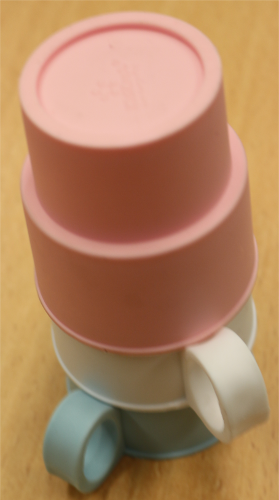}
} \\
\subfloat[\tiny{Duplo bricks : \textbf{pushable}, \textbf{liftable}}]{\label{fig:brick}
\includegraphics[width=.15\linewidth]{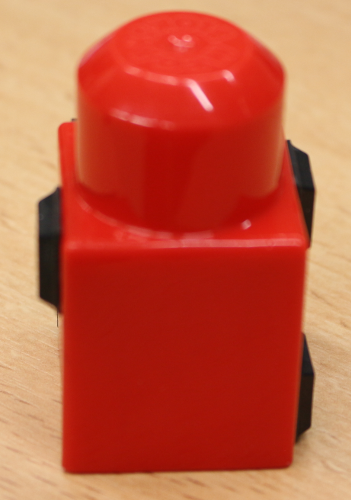}
} 
\subfloat[\tiny{Toy car : \textbf{pushable}}]{\label{fig:car}
\includegraphics[width=.2\linewidth]{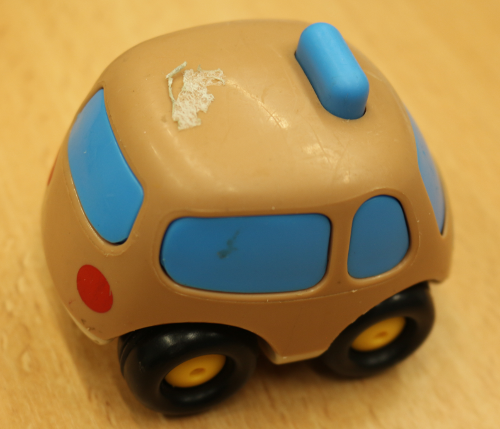}
} 
\subfloat[\tiny{Wooden cube : \textbf{pushable}, \textbf{liftable}}]{\label{fig:cube}
\includegraphics[width=.2\linewidth]{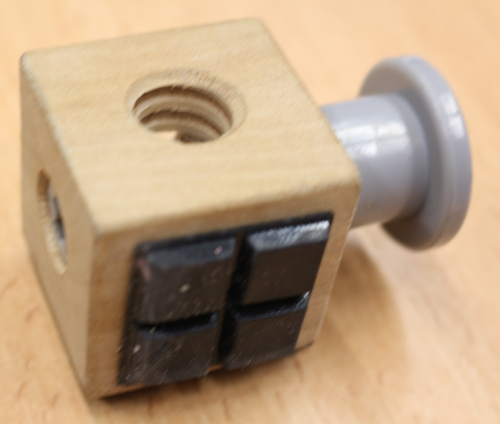}
} 
\subfloat[\tiny{Push-button : \textbf{activable}}]{\label{fig:button}
\includegraphics[width=.2\linewidth]{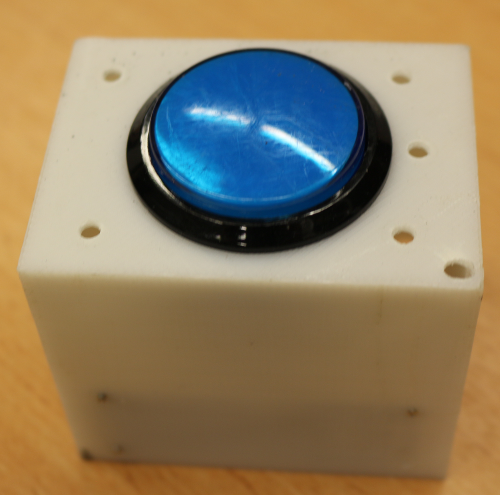}
} 
\caption{8 different types of objects used in the experiments. The affordance expected to be linked to these objects is indicated in bold.}
\label{fig:objects}
\end{figure}

Figure \ref{fig:objects} is a collection of pictures representing the objects used in the experiments: 3 bowls in a pile, 3 mugs in a pile, two different toy locomotives, Duplo bricks, two identical wooden cubes, and 5 push-buttons. Of course, the pile of bowls and mugs (see \ref{fig:bowl} and \ref{fig:mug}) can be dismantled during an experiment. The Duplo bricks are of different colors (red in the pictures \ref{fig:brick}): green, red, purple, orange, yellow. There are five push-buttons, all are visible in picture \ref{fig:aff_env}: circular blue (the one in picture \ref{fig:button}), red, yellow, green and squared green. Figure \ref{fig:objects} indicates in bold for each object its expected affordance.

\subsection{Quality measures}

To assess the performance of the trained classifier precision, recall, and accuracy are computed by following the equations \ref{eq:pra} and \ref{eq:tptn}. These measures are computed according to a ground truth. The ground truth is obtained from a snapshot of the scene without the objects that afford the studied action which corresponds to the background. For the pushable affordance, the ground truth is exact as it corresponds just to the background and the buttons. For the activable push-buttons, the ground truth is approximative because only a part of the button is activable, the colored central part (see \ref{fig:button}) while the ground truth we have set takes into account the whole white box. Thus, the performances should be slightly better than the one presented in the results section. For the liftable affordance, the ground truth is even less accurate as it corresponds to our a priori about what the robot may lift or not. 
An autonomous exploration is interesting and useful precisely when the ground truth is difficult to set. In our case, it is difficult to predict exactly the robustness of the lift according to the robot capacity and the designed lift primitive.

\begin{equation}\label{eq:pra}
\begin{split}
precision & =  \frac{tp}{tp + fp} \\
recall & =  \frac{tp}{tp + fn} \\
accuracy & = \frac{1}{2}(\frac{tp}{GT_{e}} + \frac{tn}{GT_{\overline{e}}})
\end{split}
\end{equation}
Where $tp$ is the number of true positives and $tn$ is the number of true negatives (i.e. supervoxels well classified in the class $(a,e)$ and $(a,\overline{e})$);  $fp$ are false positives, i.e. supervoxels misclassify as part of class $(a,e)$ and $fn$ are false negatives, i.e. supervoxels misclassified as part of class $(a,\overline{e})$; and $GT_{e}$ is the ground truth for parts of the environment that produced the expected effect and $GT_{\overline{e}}$ is the ground truth for parts of the environment that do not produce the expected effect. Their definitions, for N supervoxels extracted from a pointcloud, is the following:

\begin{equation} \label{eq:tptn}
\begin{split}
tp & = \sum_i^N{P(\Delta = (a,e) | W, \Theta, x_i)*(1 - \delta_i)} \\
tn & = \sum_i^N{P(\Delta = (a,\overline{e}) | W, \Theta, x_i)*\delta_i} \\
fp & = \sum_i^N{P(\Delta = (a,e) | W, \Theta, x_i)*\delta_i} \\
fn & = \sum_i^N{P(\Delta = (a,\overline{e}) | W, \Theta, x_i)*(1 - \delta_i)} \\
GT_{e} & = \sum_i^N{1 - \delta_i} \\
GT_{\overline{e}} & = \sum_i^N{\delta_i}
\end{split}
\end{equation}
Where $\delta_i$ is the Kronecker symbol equal to $1$ if the i$^{th}$ supervoxel is part of the background, and otherwise equal to $0$; $x_i$ represents the features of the i$^{th}$ supervoxel.

These measures are widely used as quality measures for supervised learning algorithm.

\section{Results}\label{sec:res}

\begin{figure}[!h]
\centering
\subfloat{
\includegraphics[width=.8\linewidth]{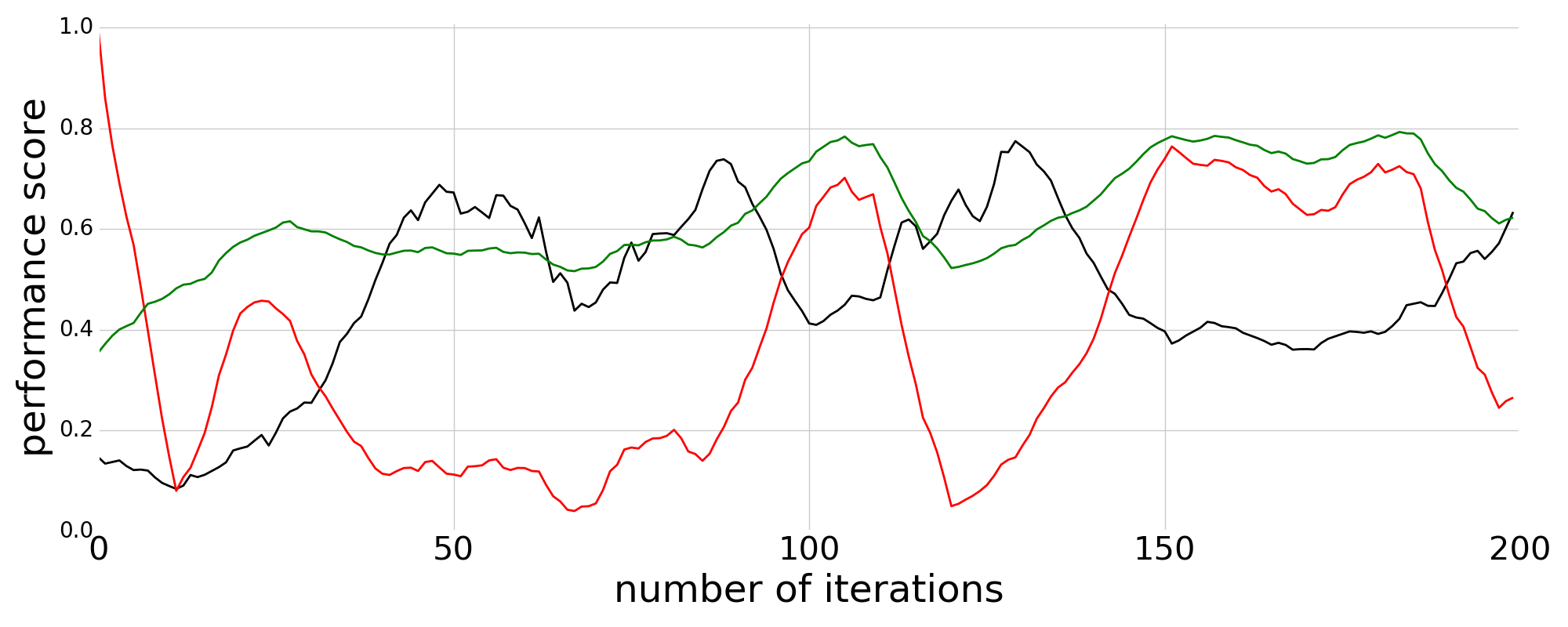}
} \\
\subfloat{
\includegraphics[width=.8\linewidth]{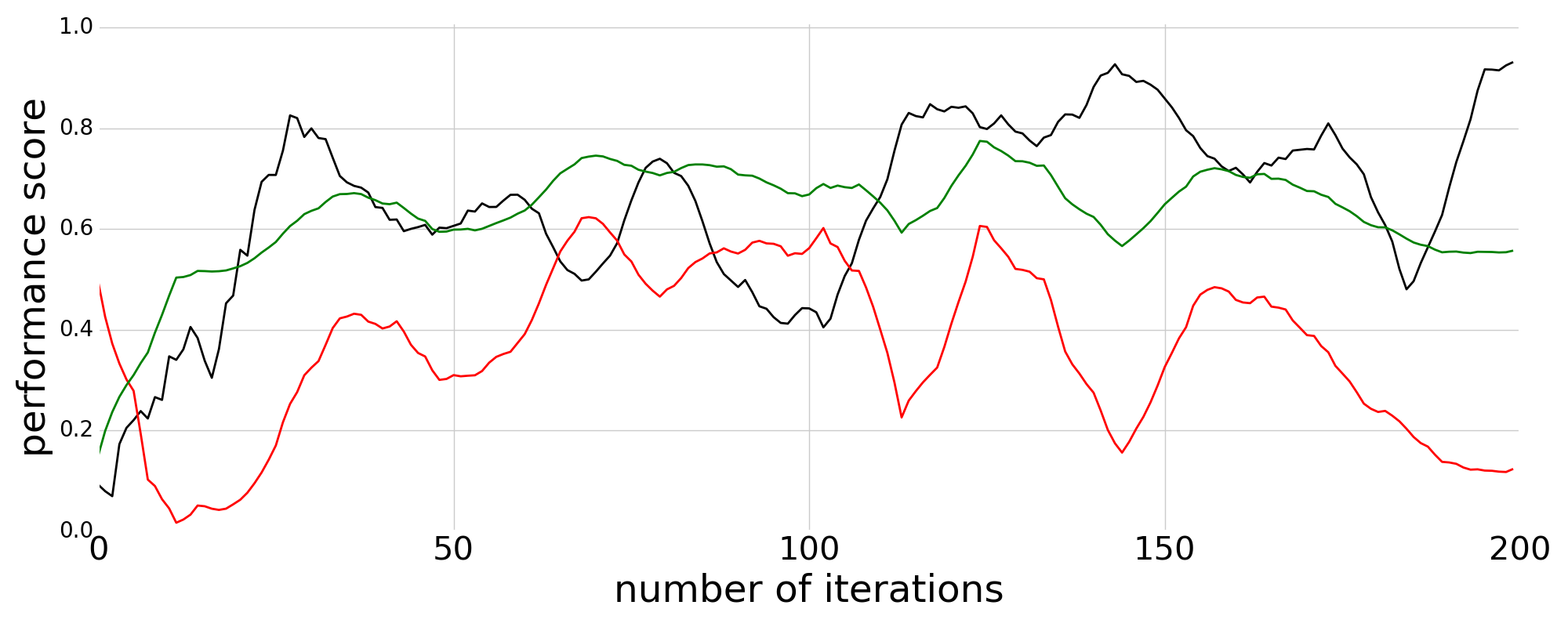}
} \\
\subfloat{
\includegraphics[width=.8\linewidth]{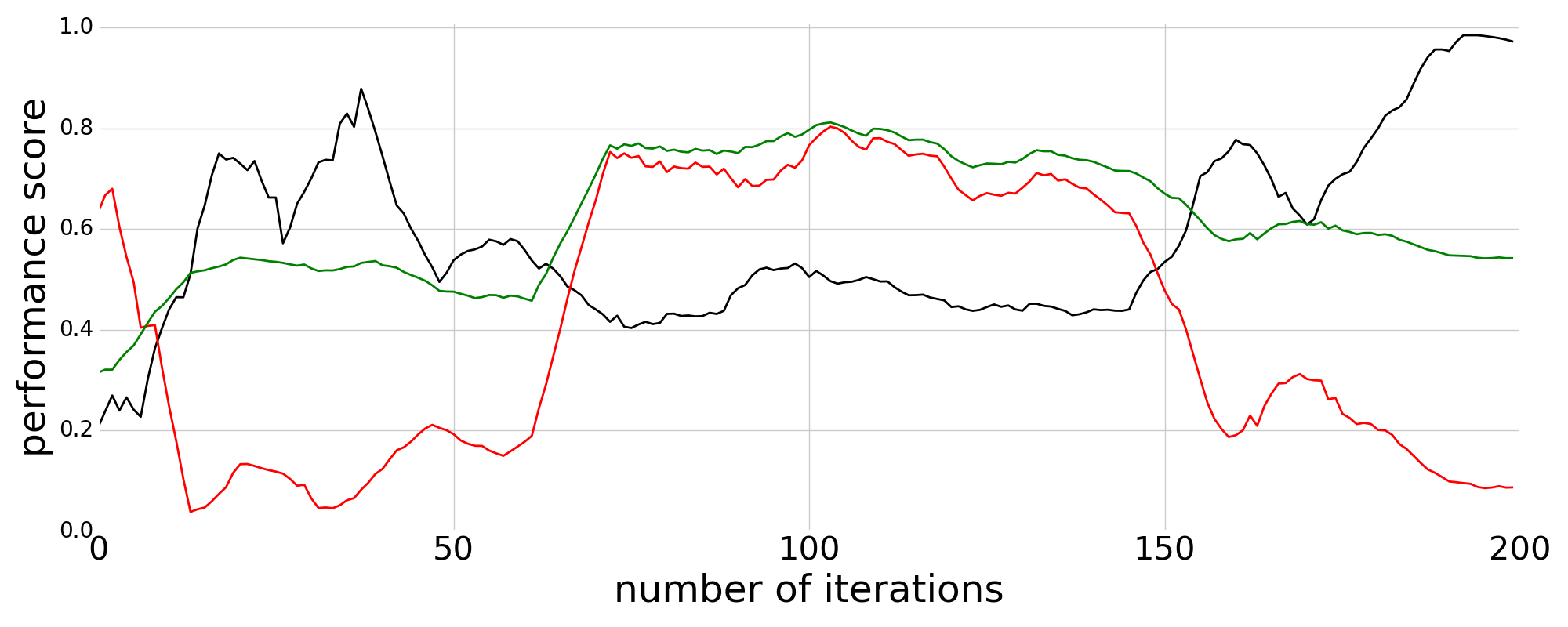}
} \\
\subfloat{
\includegraphics[width=.8\linewidth]{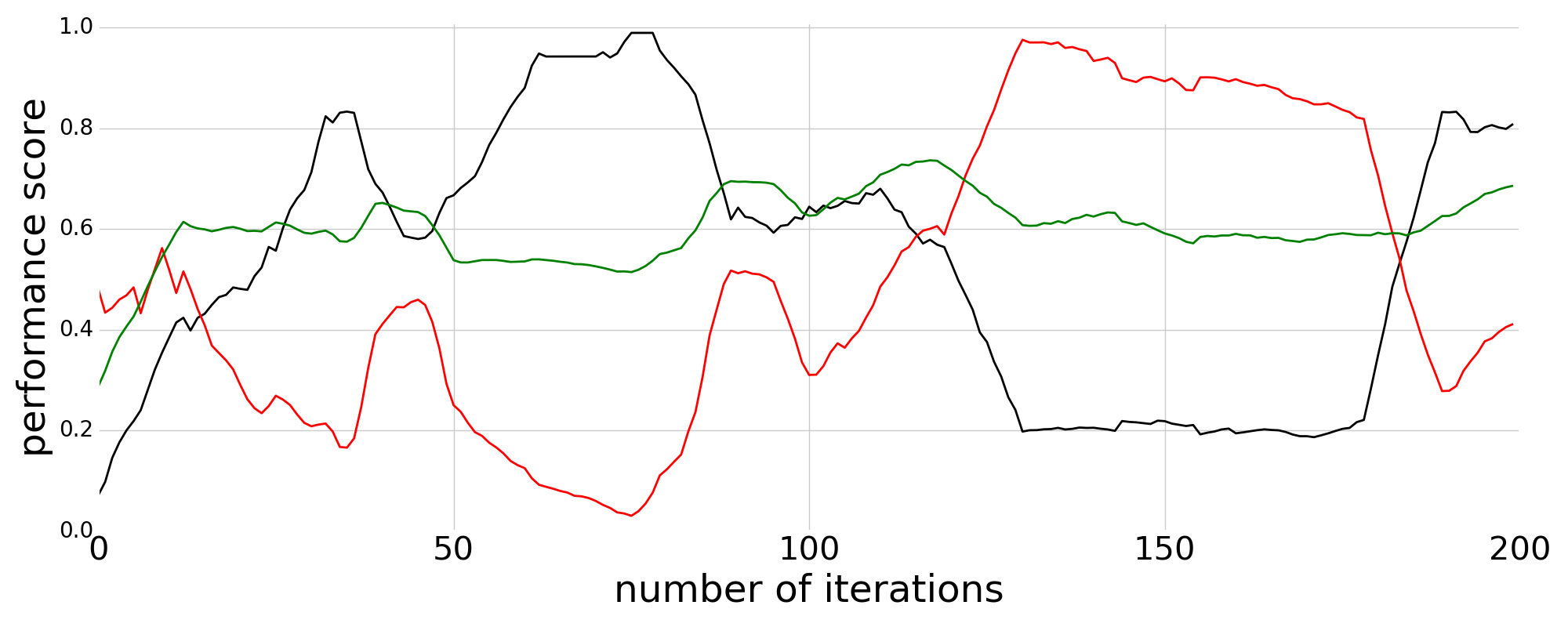}
} \\
\subfloat{
\includegraphics[width=.7\linewidth]{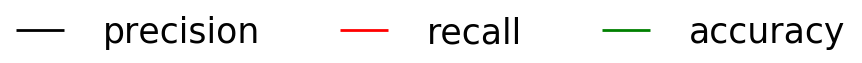}
} 
\caption{Plots of precision, recall, and accuracy for pushable affordance}
\label{fig:pushpra}
\end{figure}

For each experiment, the precision, recall, and accuracy scores of each replication are presented separately to avoid losing information. 

The precision, recall, and accuracy scores of the experiment for the pushable affordance (presented in figure \ref{fig:pushpra}) are satisfying considering the complexity of the setup. The classification quality is very different for each replication. In the first experiment (the top left part of the figure \ref{fig:pushpra}), the classifier converges only around the 150th interaction with an accuracy around 0.8, a recall varying between 0.6 and 0.8, and a low precision around 0.4. Finally, for this replication, the quality drops at the end. For the second and third experiments (the top right and the bottom left parts of the figure \ref{fig:pushpra}) the classifier converges around the 60th interaction. For the second replication, the accuracy, recall, and precision are not stable and the classifier starts diverging after the 100th interaction. The classifier, of the third replication, converges to an accuracy and a recall around 0.8 and a precision between 0.4 and 0.5 and stays stable. But it diverges after the 150th interactions. For the last replication (the bottom right part of the figure \ref{fig:pushpra}), it is difficult to isolate a period of convergence of the classifier. The classification quality of this experiment is very unstable.

\begin{figure}[!h]
\centering
\subfloat{
\includegraphics[width=.46\linewidth]{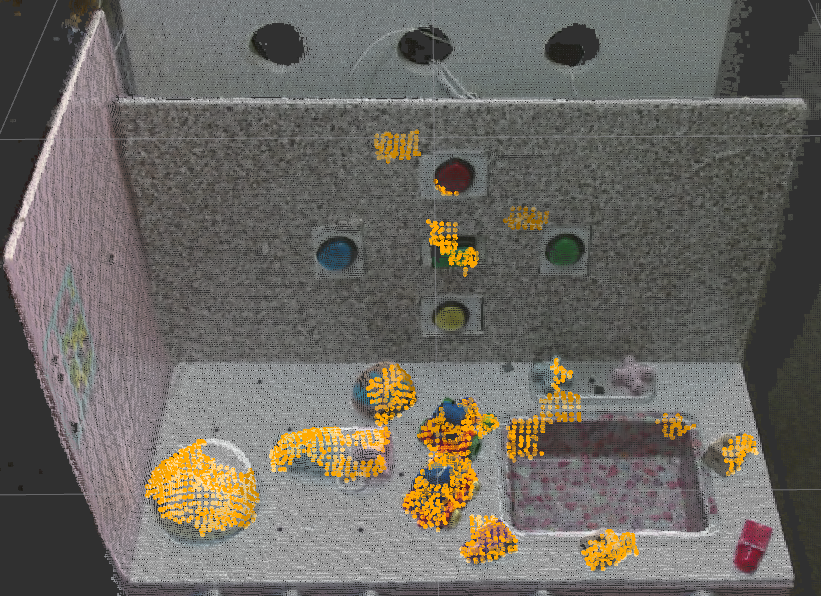}
}
\subfloat{
\includegraphics[width=.46\linewidth]{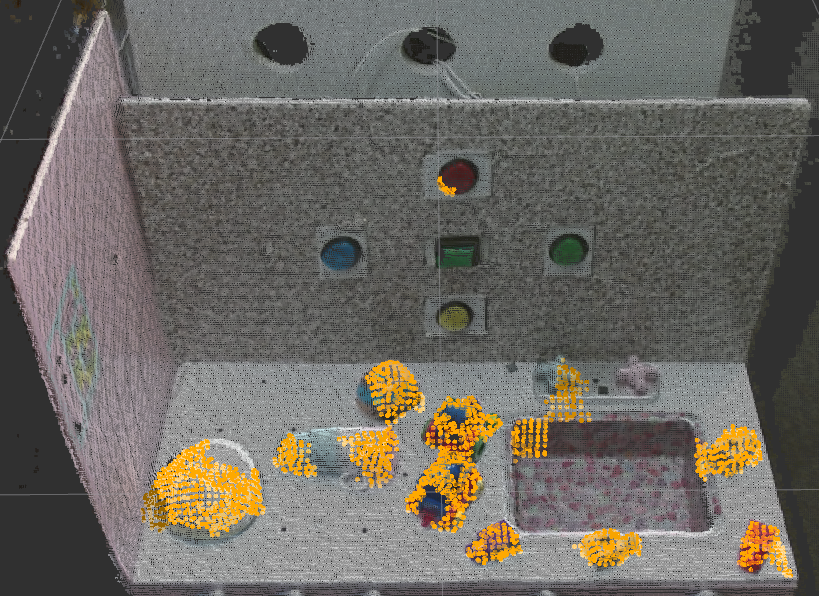}
} \\
\subfloat{
\includegraphics[width=.46\linewidth]{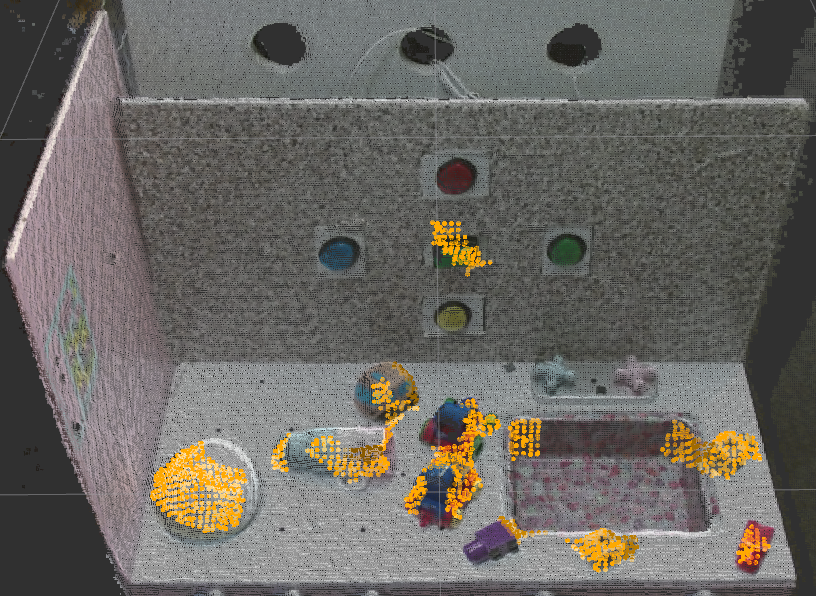}
}
\subfloat{
\includegraphics[width=.46\linewidth]{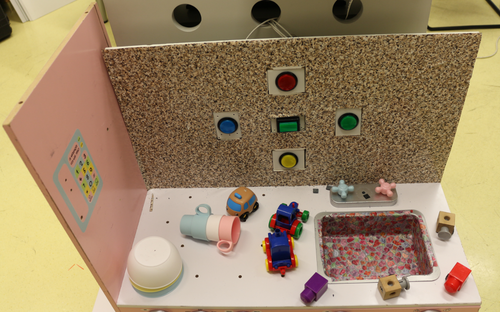}
}
\caption{Three push relevance maps extracted from the same scene and with the same classifier on a video stream. The differences between the three maps is due to the extraction of the supervoxels which produces a different segmentation at each frame. The bottom picture represents the environment from which the relevance maps have been extracted.}
\label{fig:rmvar}
\end{figure}

 For all the replications, the quality of classification diverges at the end. The divergence is probably due to mislabeled samples and to splitting or merging components which were not suitable to represent the data. The instability of the classification quality, clearly visible in the second replications, is due to the inconsistency of the supervoxel segmentation when extracted on a video stream as shown in figure \ref{fig:rmvar}. 

The figure \ref{fig:rmvar} shows three pictures representing push relevance maps. These relevance maps have been extracted with the same classifier on the same static scene on a video stream. The variability of the relevance map over these three images are due to the extraction of the supervoxels which produces a different segmentation at each frame. The variability of the segmentation is due to the noise of the depth stream. The higher the noise, the higher the variability is. On these pictures, the toy locomotives and the button are the noisiest areas. On these areas, the geometrical features can change a lot, which is due to the variation in the shape of the supervoxels.  

\begin{figure}[!h]
\centering
\subfloat{
\includegraphics[width=.8\linewidth]{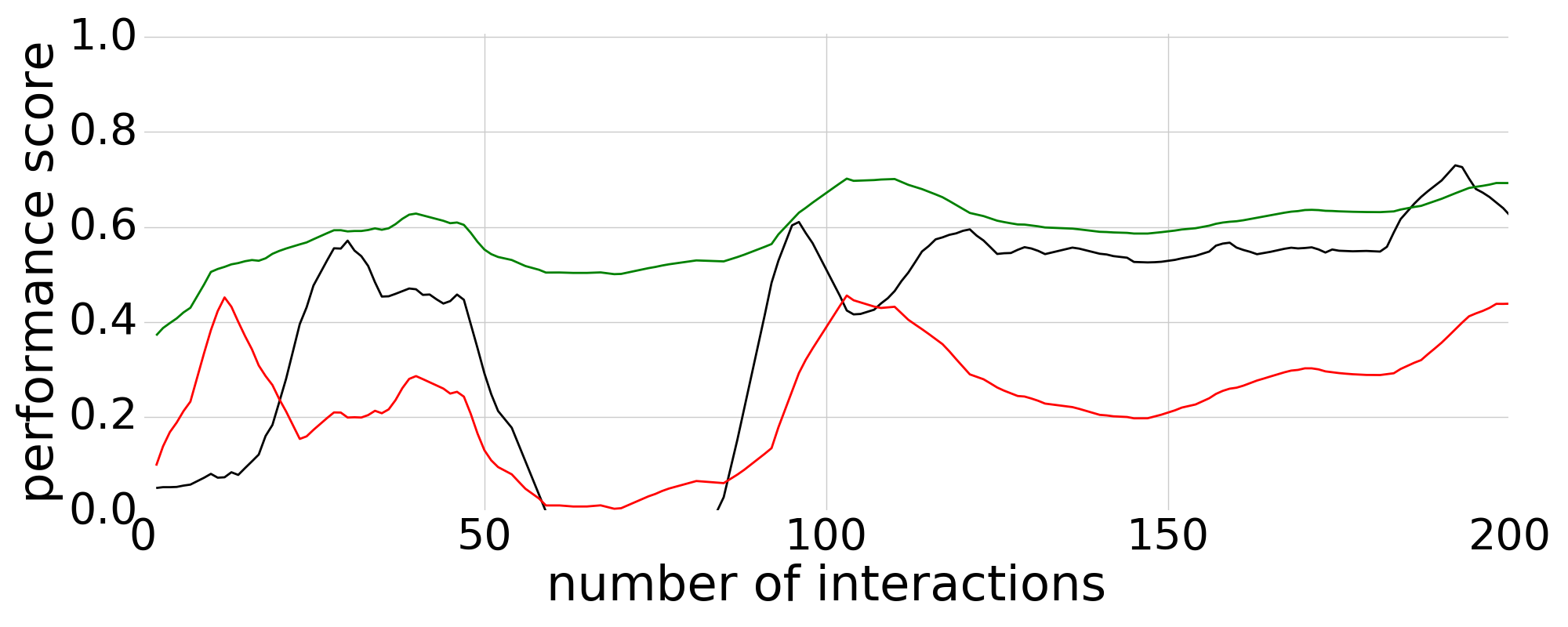}
}\\
\subfloat{
\includegraphics[width=.8\linewidth]{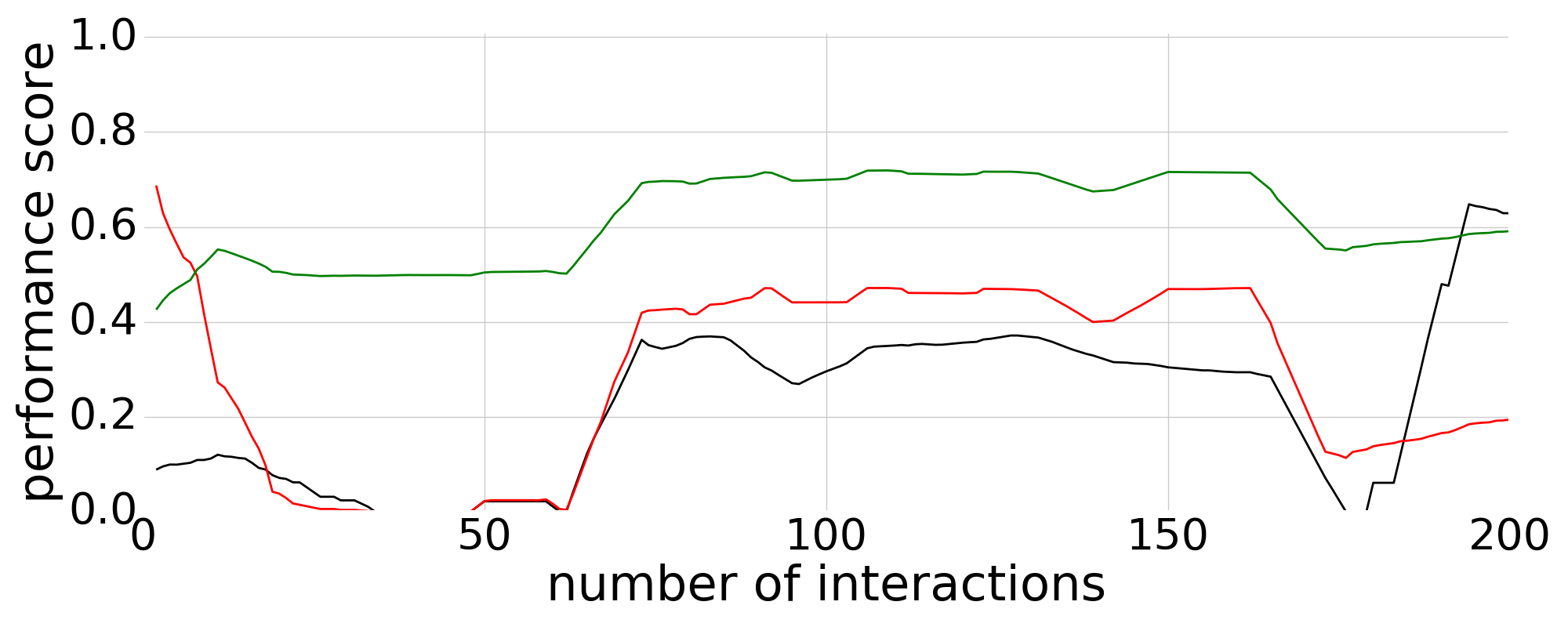}
} \\
\subfloat{
\includegraphics[width=.8\linewidth]{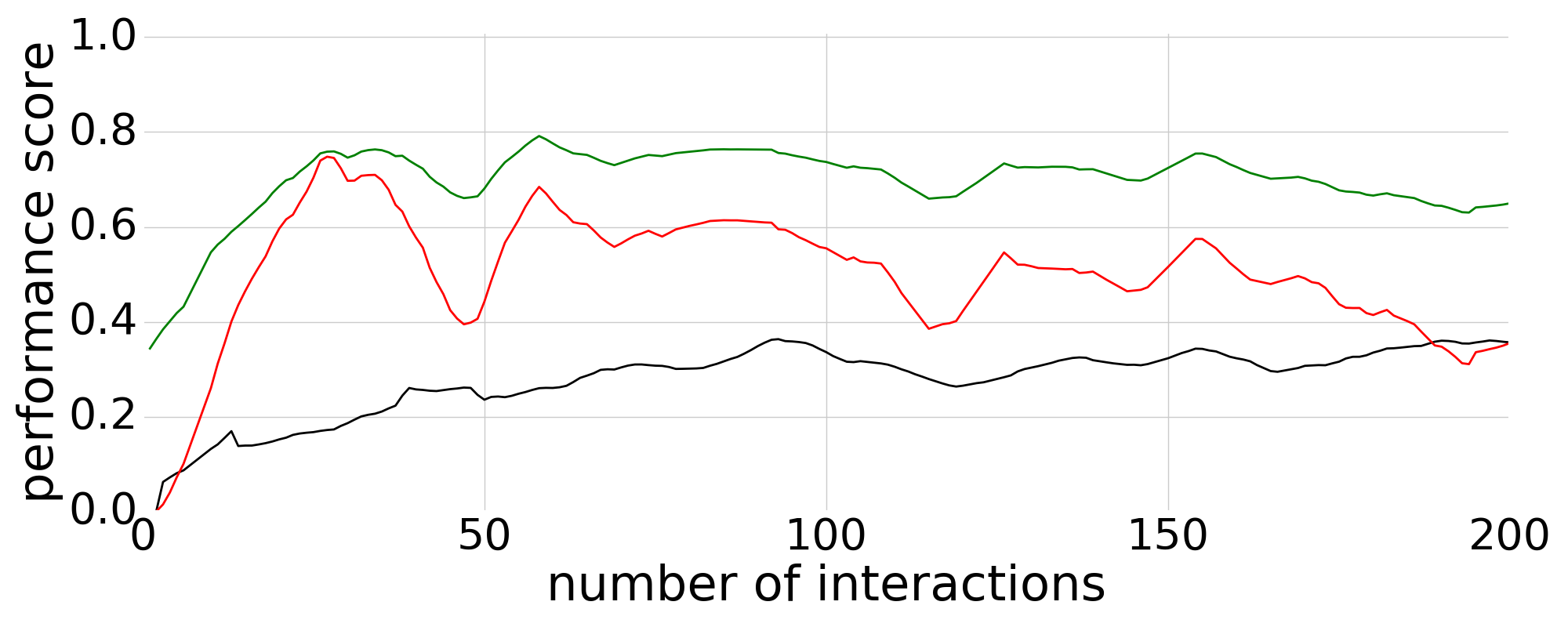}
}\\
\subfloat{
\includegraphics[width=.8\linewidth]{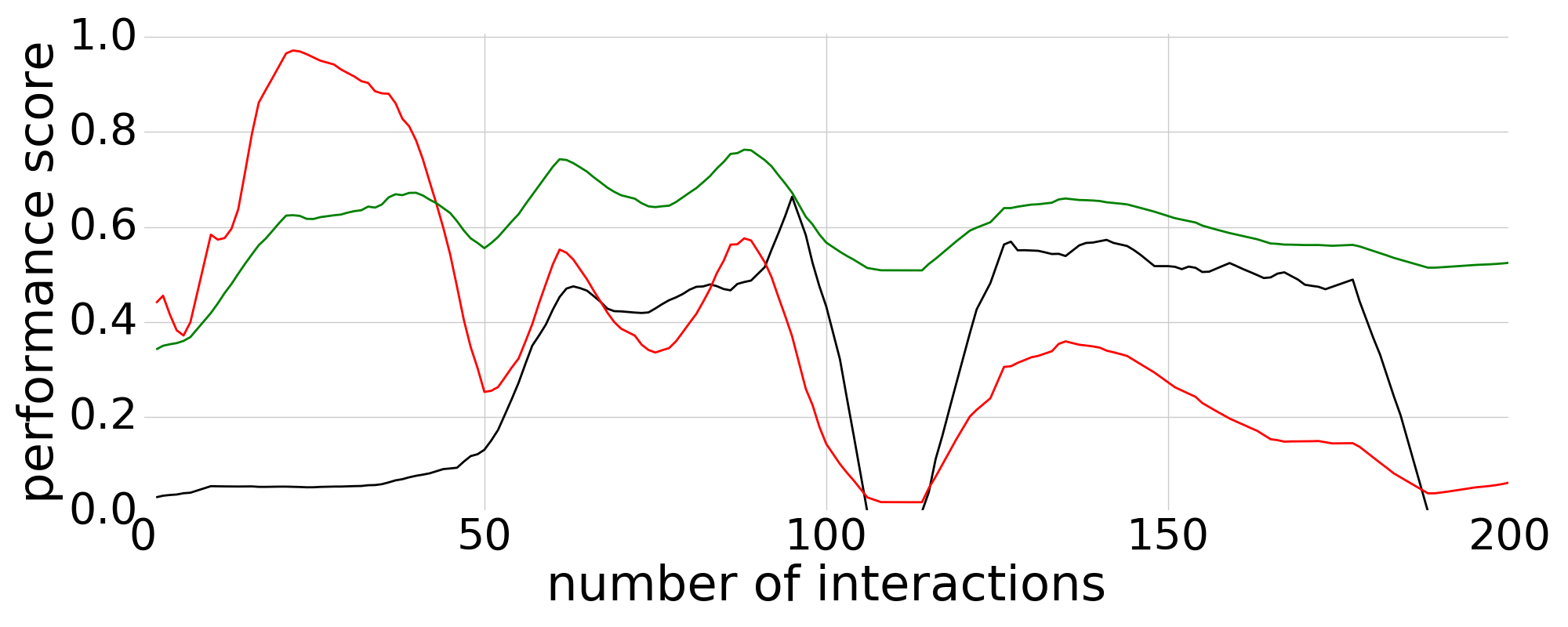}
} \\
\subfloat{
\includegraphics[width=.7\linewidth]{img/legend_pra}
} 
\caption{Plots of precision, recall, and accuracy for activable push-buttons}
\label{fig:buttonpra}
\end{figure}

The precision, recall, and accuracy scores of the experiment with the push-buttons are shown in the figure \ref{fig:buttonpra}. In this experiment, the replications give also different results. For the first replication (the top left part of the figure \ref{fig:buttonpra}), the classifier converges around the 80th interaction and keep the quality of classification steady around a value of 0.6 for the accuracy and the precision, a value of 0.5 for the recall. For the second replication (top right of the figure \ref{fig:buttonpra}), the classifier converges around the 75th interactions with an accuracy around 0.7, a recall around 0.5 and a precision under 0.4, but this replication starts to diverge around the 160th interaction. For the third replication (the bottom left part of the figure \ref{fig:buttonpra}), the classifier converges quickly to a value between 0.7 and 0.8 for the accuracy, around 0.6 for the recall while the precision increases slowly during all the replication. The accuracy and the recall slowly decrease after the 100th interaction. Finally, the last replication (bottom right of the figure \ref{fig:buttonpra}) presents poor results. The classifier converges first between the 50th and 100th interaction, then diverges, and then converges again to a low classification quality, before finally diverging.

Overall, the classification is more stable for this experiment than for the experiments with the pushable affordance. The main difficulty in this experiment is that the buttons represent a small area. The size of the actual pushable area is even smaller, about the size of a supervoxel. This introduces noise on the extracted features. A solution may be to reduce the size of the supervoxels, but if a supervoxel contains too few points, the features could be inconsistent. Moreover, this reduced size creates a strong requirement in terms of the accuracy of the action primitive to prevent mislabeling.

\begin{figure}[!h]
\centering
\subfloat{
\includegraphics[width=.8\linewidth]{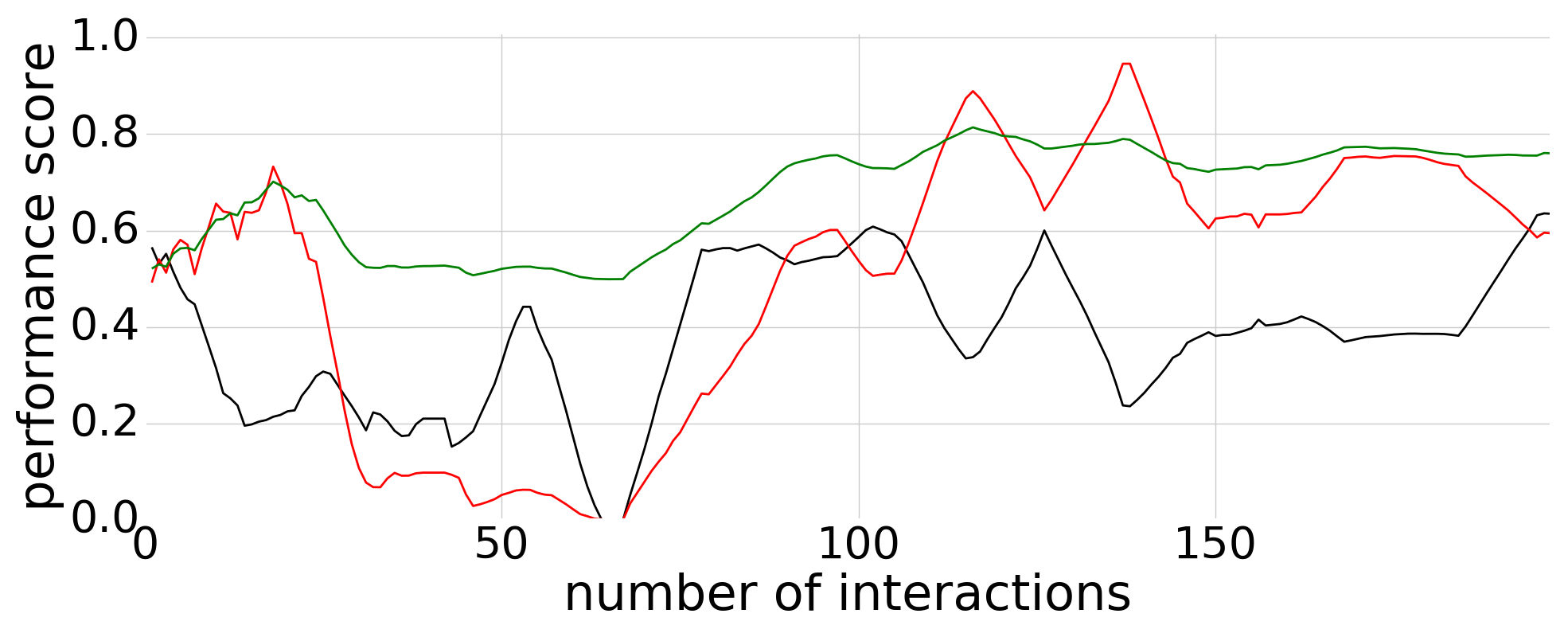}
} \\
\subfloat{
\includegraphics[width=.8\linewidth]{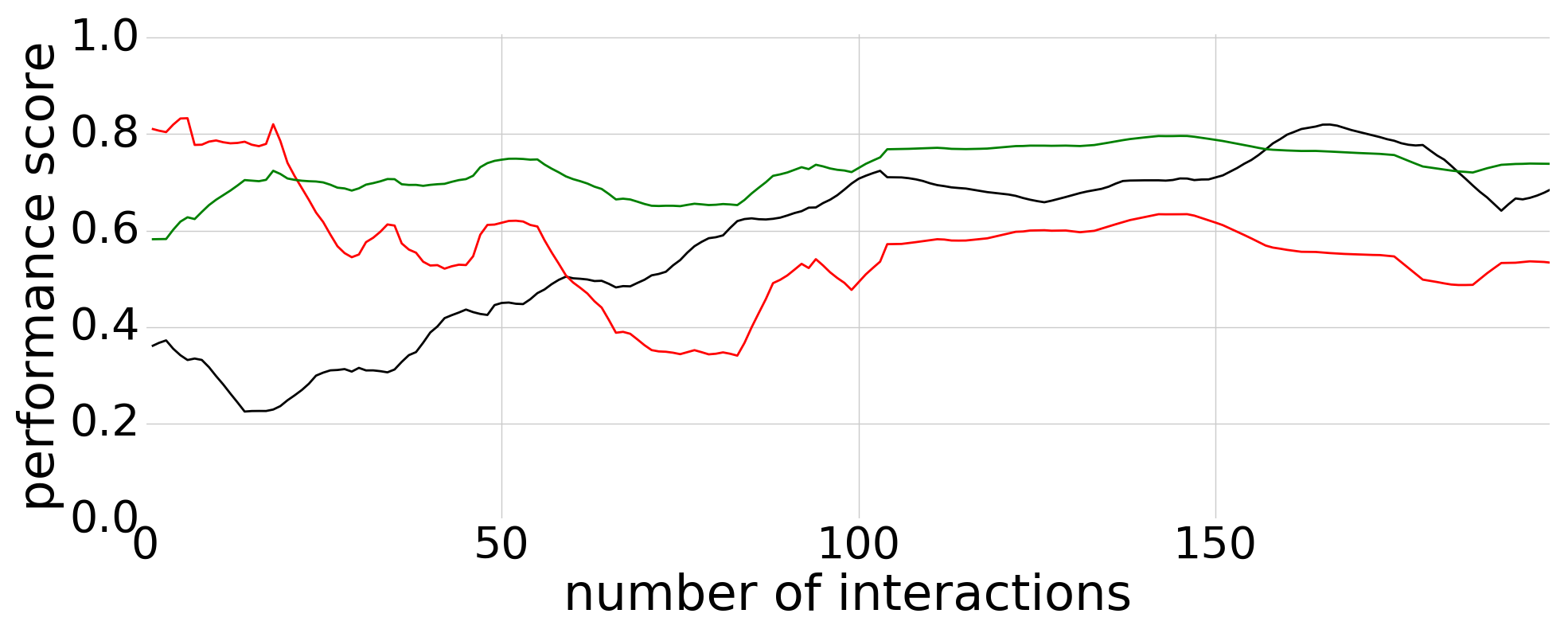}
} \\
\subfloat{
\includegraphics[width=.8\linewidth]{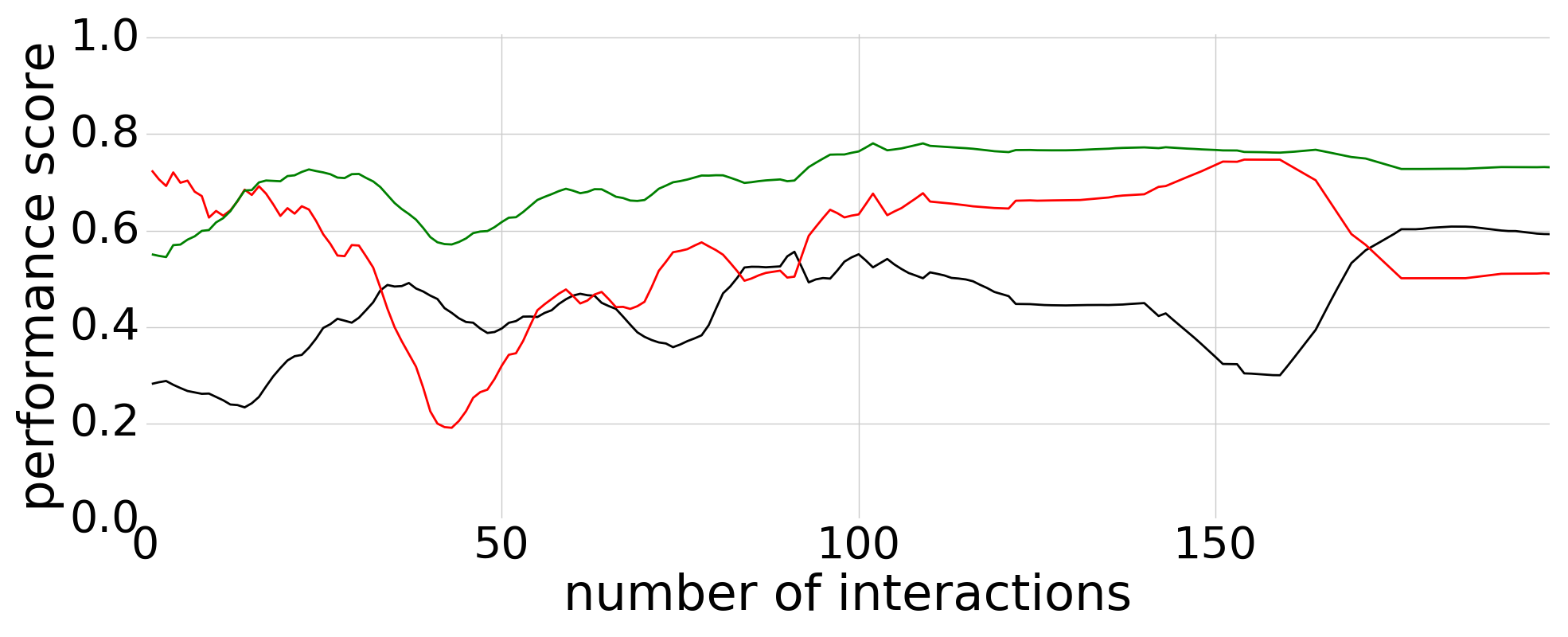}
} \\
\subfloat{
\includegraphics[width=.8\linewidth]{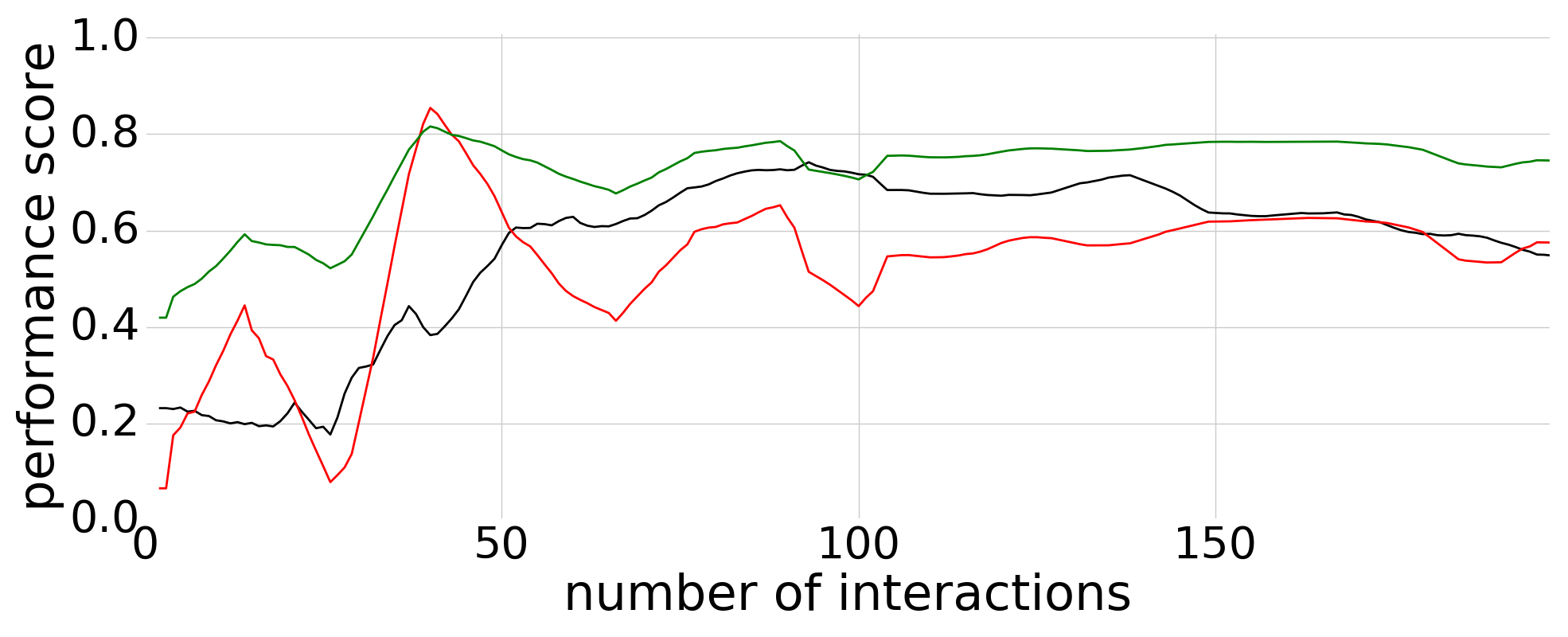}
} \\
\subfloat{
\includegraphics[width=.7\linewidth]{img/legend_pra}
} 
\caption{Plots of precision, recall, and accuracy for liftable affordance}
\label{fig:liftpra}
\end{figure}

Figure \ref{fig:liftpra} represents the performances monitored during the experiment conducted for the liftable affordances. For the first and the third replications (the left part of figure \ref{fig:liftpra}), the quality scores have similar shapes, the convergence is reached around the 100th interaction with a low precision and an accuracy, and a recall between 0.7 and 0.8. For the first replication, the recall, and precision are unstable between the 100th and the 150th interactions. In both, the recall and precision cross themselves to have a higher precision than recall which can be seen with a light decrease of the accuracy.  For the second and fourth replications (the right part of figure \ref{fig:liftpra}), the classifier converges after the 100th interaction, with an accuracy around 0.8, a recall around 0.6, and a higher precision around 0.7. Unlike the two previous experiments (pushable and push-button), the classification quality does not seem to diverge at the end of the experiment, except for the forth replication for which the precision decreases slowly after the 150th interaction.

As in the previous experiment, this experiment gives stable results. The low precision, observed on the first and third replications, is probably due to the inaccuracy of the ground truth. Finally, the stability of the convergence may be due to the use of a push relevance map to filter the classification which does not change during the experiment. However, this does not explain entirely the absence of divergence.

\begin{figure}[!h]
\centering
\subfloat{
\includegraphics[width=.9\linewidth]{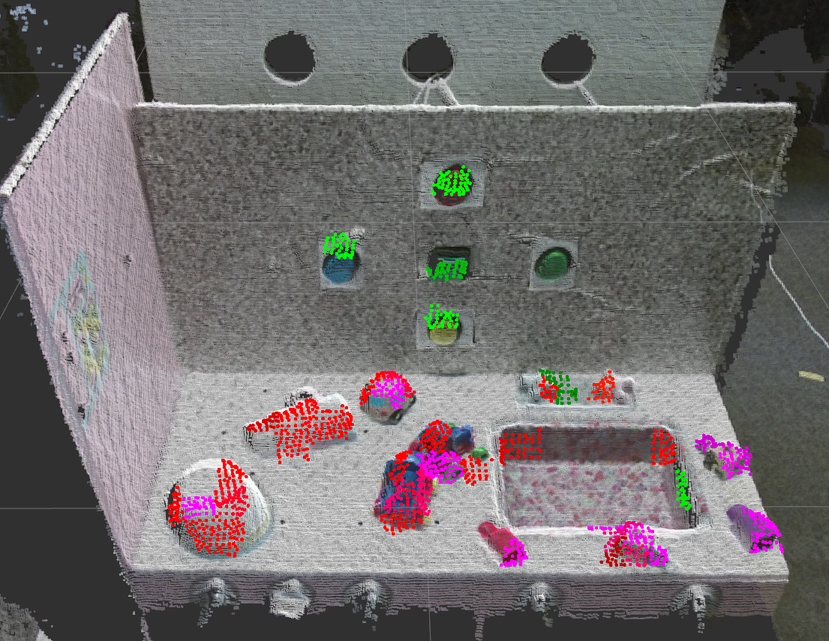}
}\\
\subfloat{
\includegraphics[width=.5\linewidth]{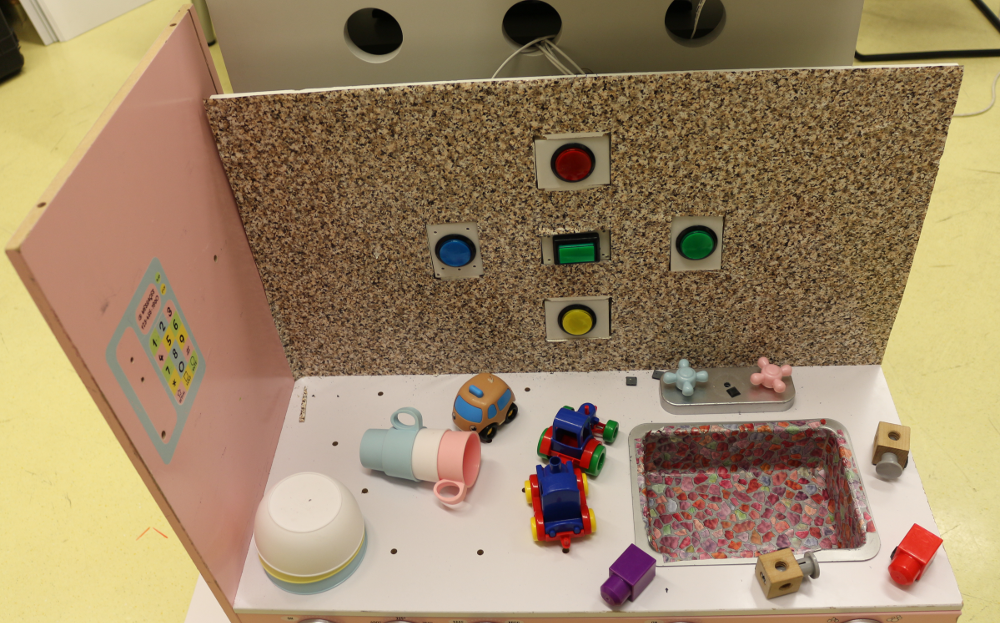}
}  
\caption{Affordances map of liftable activable push-buttons and pushable affordances. Colored areas indicate areas classified with a probability above 0.5, in red to afford the push primitive, in purple to afford the lift primitive and in green to be an activable push-buttons. The bottom picture represents the environment on which the affordances map has been extracted.} 
\label{fig:aff_map}
\end{figure}

Figure \ref{fig:aff_map} represents an affordances map obtained thanks to the experiments described above. This map represents the areas categorized as pushable buttons in green, as liftable objects in yellow, and as pushable objects in orange. It was obtained by selecting the best performing classifier among the experiments and at the best moment inside a replication. Only supervoxels of both relevance maps with a probability equal or higher of 0.5 are displayed. 

An interesting property in this affordance map, is the low overlap between the parts predicted to be pushable and to be a push-button. Also, as expected, the pile of bowl is detected as only pushable. The other objects are predicted as pushable and liftable or partially liftable. This affordances map is a proof of concept of what can be obtained with the proposed approach. For each experiment, more replications are required for a better assessment of the method robustness. The instability of the classifier needs to be dealt with for this approach to be more reliable.

The source code use to produce this results can be found on github\footnote{\url{https://github.com/robotsthatdream/wave1_relevance_map}}.

\section{Discussion and Future work}\label{sec:disc}

The experiments described in this article provide a proof of concept of how the proposed approach can be used to learn an affordances map. Although the results have shown a large variability over the four replications done for each affordance, relevance maps have been produced and combined into a meaningful affordances map. The relevance maps of the push-button and of pushable affordances do not overlap, which shows the capacity of the classifier to learn different concepts. The classifier is also able to refine a concept as it is shown in the experiment with the liftable affordance.

The robustness and stability of the method can still be improved in the complex environments it was tested on here, as the classifiers trained for the pushable and activable push-button affordances diverge on the four replications. The poor accuracy of the button pushing and object lifting primitives probably plays a significant role in this instability. As the generated affordances map represents the ability of action primitives to generate expected effects on each part of the environment, the precision and success rate of those primitives is critical. Further work on creating more elaborate, more reliable action primitives is thus expected to significantly improve the system's performance.

Another approach that could be pursued in a future study is to add "reachability" affordances: for each action primitive, a "reachabililty" relevance map could be learned by   testing the areas where the robot can apply the action primitive. Then, the classification for the affordances linked to the used primitive will be filtered by the associated "reachability" map. However, to learn "reachability", the proposed framework will have to be extended, as spatial or proprioceptive information should be used as features, which differs from the information currently used in our method. For instance, if spatial information is combined with FPFH and color histograms, the generalization in term of the spatial position of the objects will be lost. Moreover, the CMMs classifier may not be adapted to learn from spatial or proprioceptive information. Another classifier could be used with other features, while keeping the same architecture.

\section{Conclusion}\label{sec:con}

In this study, a method is proposed to learn different preceptual maps called relevance maps relative to specific affordances. The framework is modular and thus permits to learn relevance maps relative to different affordances. In this paper, as proof of concept, experiments have been conducted to learn relevance maps relative to pushable, liftable and activable push-button. Then, by combining these maps, a new perceptual map is obtained, called affordances map. We call it this way as it allows the robot to perceive the environment through its possible actions.

\section*{Acknowledgement}
This work is supported by the DREAM project \footnote{\url{http://www.robotsthatdream.eu}} through the European Unions Horizon 2020 research and innovation program under grant agreement No 640891. This work has been partially sponsored by the French government research program Investissements d'avenir through the Robotex Equipment of Excellence (ANR-10-EQPX-44).

\bibliographystyle{IEEEtran}
\bibliography{bib}

\end{document}